\documentclass[10pt,journal,cspaper,compsoc]{IEEEtran}
\ifCLASSOPTIONcompsoc
\else
\fi
\ifCLASSINFOpdf
\else
\fi
\usepackage[cmex10]{amsmath}
\usepackage{array}

\usepackage[tight,scriptsize]{subfigure}
\usepackage{url}

\usepackage{graphicx}
\usepackage{booktabs}
\usepackage{multirow}
\usepackage{float}
\usepackage{dsfont}
\usepackage{amssymb}
\usepackage[boxed,vlined]{algorithm2e}
\usepackage{cite}
\usepackage{xcolor}
\usepackage{threeparttable}

\newcommand{\etal}{\mbox{\emph{et al.\ }}}

\newcommand{\ie}{\mbox{\emph{i.e.,\ }}}
\newcommand{\eg}{\mbox{\emph{e.g.,\ }}}
\newcommand{\tabincell}[2]{\begin{tabular}{@{}#1@{}}#2\end{tabular}}

\begin{document}
%
\title{Heterogeneous Face Attribute Estimation: \\A Deep Multi-Task Learning Approach}
%
%
%
%

\author{Hu~Han,~\IEEEmembership{Member,~IEEE,}
        Anil~K.~Jain,~\IEEEmembership{Fellow,~IEEE,}
        Fang~Wang,\\
        Shiguang~Shan,~\IEEEmembership{Senior Member,~IEEE} 
        and~Xilin~Chen,~\IEEEmembership{Fellow,~IEEE}
\IEEEcompsocitemizethanks{\IEEEcompsocthanksitem Hu Han, Fang Wang, Shiguang Shan, and Xilin Chen are with the Key Laboratory of Intelligent Information Processing of Chinese Academy of Sciences (CAS), Institute of Computing Technology, CAS, Beijing, 100190, China, and the University of Chinese Academy of Sciences, Beijing 100049, China. Shiguang Shan is also with the CAS Center for Excellence in Brain Science and Intelligence Technology.
\protect\\
Anil K. Jain is with the Department of Computer Science and Engineering, Michigan State University, East Lansing, MI 48824, USA. \protect\\

E-mail: \{hanhu, sgshan, xlchen\}@ict.ac.cn; jain@cse.msu.edu; fang.wang14@vipl.ict.cn
}
\thanks{Early versions of this work appeared in the MSU technical report (MSU-CSE-14-5), 2014 \cite{HanTR14}, and the Proceedings of the 12th IEEE International Conference on Automatic Face and Gesture Recognition (FG), 2017 \cite{WangFG17}.}
}

%


%
%

\markboth{}%
{Han \MakeLowercase{\textit{et al.}}: Heterogeneous Face Attribute Estimation: A Deep Multi-Task Learning Approach}
%


\IEEEcompsoctitleabstractindextext{%
\begin{abstract}

Face attribute estimation has many potential applications in video surveillance, face retrieval, and social media.
While a number of methods have been proposed for face attribute estimation, most of them did not explicitly consider the attribute correlation and heterogeneity (\eg ordinal vs. nominal and holistic vs. local) during feature representation learning.
In this paper, we present a \emph{Deep Multi-Task Learning (DMTL)} approach to jointly estimate multiple heterogeneous attributes from a single face image.
In DMTL, we tackle attribute correlation and heterogeneity with convolutional neural networks (CNNs) consisting of shared feature learning for all the attributes, and category-specific feature learning for heterogeneous attributes.
We also introduce an unconstrained face database (LFW+), an extension of public-domain LFW, with heterogeneous demographic attributes (age, gender, and race) obtained via crowdsourcing.
Experimental results on benchmarks with multiple face attributes (MORPH II, LFW+, CelebA, LFWA, and FotW) show that the proposed approach has superior performance compared to state of the art.
Finally, evaluations on a public-domain face database (LAP) with a single attribute show that the proposed approach has excellent generalization ability.

\end{abstract}

}

\maketitle

\IEEEdisplaynotcompsoctitleabstractindextext

%
\IEEEpeerreviewmaketitle

\section{Introduction}

\IEEEPARstart{H}{uman} face portrays important cues for social interaction, providing a wide variety of salient information, including the person's identity, demographic (age, gender, and race), hair style, clothing, etc.
Over the past $50$ years, significant advances have been made in extracting discriminative features in a face image to determine the subject's identity \cite{FRHandbook}.
In more recent years, several applications have emerged that make use of face attributes, from demographic attributes (\eg age, gender, and race) to descriptive visual attributes (\eg clothing and hair style).
These applications include (i) \emph{video surveillance} \cite{vaquero2009attribute} \cite{kim2012attribute}, \eg automatic detection of persons with sunglasses or mask observed at unusual hours or in unusual places; (ii) \emph{face retrieval} \cite{WuTPAMI11} \cite{KumarTPAMI11} \cite{xia2012toward}, \eg automatic filtering of a face database to find person(s) of interest with given attributes; and (ii) \emph{social media} \cite{qi2012exploring} \cite{qi2009learning}, \eg automatic recommendation of hair styles or makeups.

\begin{figure}[t]
\centering
\includegraphics[width=0.99\linewidth,height=0.5\linewidth]{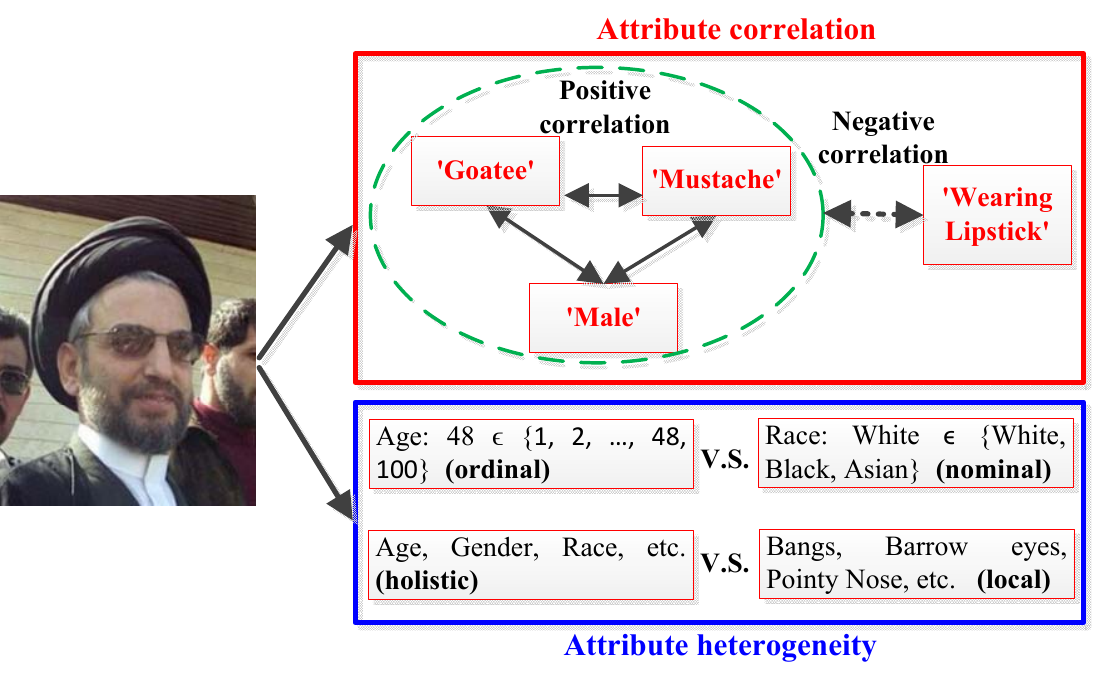}
\vspace{-6mm}
\caption[AttributeCorrHete]{Individual face attributes have both correlation and heterogeneity. While attribute correlation can be utilized to improve the robustness of attribute estimation, attribute heterogeneity should also be tackled by designing appropriate prediction models.}
\label{Fig.TraitInFaceImage}
\vspace{-6mm}
\end{figure}

\begin{figure*}[t]
\centering
\includegraphics[width=.9\linewidth]{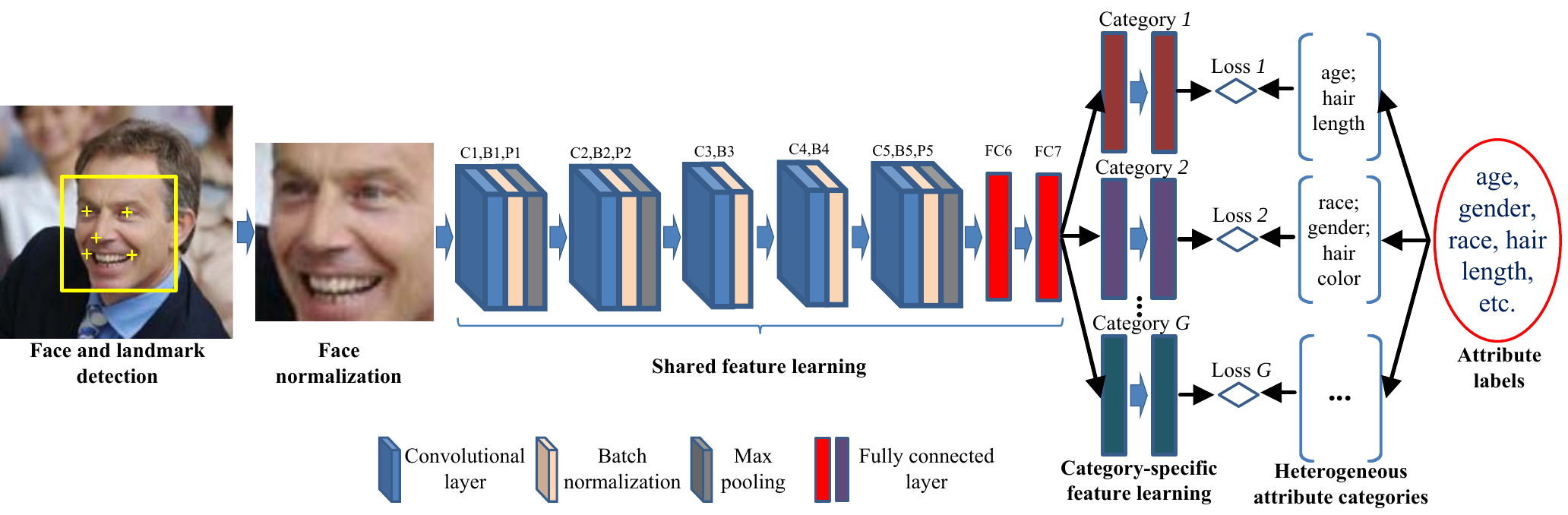}
\vspace{-3mm}
\caption{Overview of the proposed deep multi-task learning (DMTL) network consisting of an early-stage shared feature learning for all the attributes, followed by category-specific feature learning for heterogeneous attribute categories. We use a modified AlexNet \cite{KrizhevskyNIPS12} with a batch normalization (BN) layer inserted after each Conv. layer for shared feature learning. The subnetworks are used to fine-tune the shared features towards the optimal estimation of individual heterogeneous attributes, \eg nominal vs. ordinal and holistic vs. local.}
\label{Fig.DMTL}
\vspace{-6mm}
\end{figure*}

Despite recent progresses in face attribute prediction \cite{FuPAMI10,NiTMM11,KumarTPAMI11,GengPAMI13,EidingerTIFS14,HanTPAMI15,RotheIJCV16,SunTPAMI17}, most prior work is limited to estimating a single face attribute (\eg age), or learning a separate model for each face attribute.
To address these limitations, attempts have been made to develop new approaches that explore attribute correlation for \emph{joint} estimation of multiple face attributes \cite{GuoIVC14,YiACCV14,LuoICCV13,ZhangTPAMI16,LiuICCV15}.
Even these methods have some serious limitations.
For example, approaches in \cite{GuoIVC14,YiACCV14,ZhangTPAMI16} used the same features for estimating all the attributes without considering the attribute heterogeneity.
The sum-product network (SPN) adopted in \cite{LuoICCV13} for modeling attribute correlations may not be feasible because of the exponentially growing number of attribute group combinations.
The cascade network in \cite{LiuICCV15} also required learning a separate Support Vector Machine (SVM) classifier for each face attribute, and is not an end-to-end learning approach.

Figure \ref{Fig.TraitInFaceImage} shows that a face image portrays a wide variety of attributes, which are both \emph{correlated} and \emph{heterogeneous}.
Attribute correlation can be either positive or negative.
For example, a person with goatee and mustache is more likely to be a male, and is less likely to wear lipstick.
Meanwhile, individual attributes can be heterogeneous in terms of \emph{data type and scale} \cite{Jain88}, and \emph{semantic meaning} \cite{SamangooeiBTAS08}.
While attributes like age and hair length are ordinal, attributes like gender and race are nominal; these two categories of attributes are heterogeneous in terms of data type and scale.
Similarly, while attributes such as age, gender, and race describe the characteristics of the whole face, attributes such as pointy nose and big lips, mainly describe the characteristics of local facial components; these two categories of attributes are heterogeneous in terms of semantic meaning.
Such attribute correlation and heterogeneity should be considered in designing face attribute estimation models.

Though a number of commercial systems (\eg Affectiva, Emotient, Face++, and Microsoft)\footnotemark[1]\footnotetext[1]{Affectiva: \url{www.affectiva.com}; Emotient: \url{www.emotient.com}; Face++: \url{www.faceplusplus.com}; Microsoft: \url{www.how-old.net}} provide estimates of attributes like age, gender and expression, the underlying algorithms used in commercial systems are proprietary; in addition, the databases used by these commercial engines are not (or no longer) available to the research community.
Robust estimation of a large number of heterogeneous attributes from a face image remains a challenging problem, particularly under unconstrained sensing and uncooperative subject scenarios.

\subsection{Proposed Approach}
\label{Sec.ProposedApproach}

We present a Deep Multi-Task Learning (DMTL) approach to jointly estimate multiple heterogeneous attributes from a single face image.
The proposed approach is motivated by recent advances in face attribute prediction, but takes into account both attribute correlation and attribute heterogeneity in a single convolutional neural network (CNN).
The proposed DMTL consists of an early-stage shared feature learning for all the attributes, followed by category-specific feature learning for heterogeneous attribute categories (see Fig.~\ref{Fig.DMTL}).
The shared feature learning naturally exploits the relationship between tasks to achieve robust and discriminative feature representation.
The category-specific feature learning aims at fine-tuning the shared features towards the optimal estimation of each heterogeneous attribute category.
Given the effective shared feature learning and category-specific feature learning, the proposed DMTL achieves promising attribute estimation accuracy while retaining low computational cost, making it of value in many face recognition applications.

The main contributions of this paper include: (i) an efficient multi-task learning (MTL) method for joint estimation of a large number of face attributes; (ii) modeling both attribute correlation and attribute heterogeneity in a single network; (iii) studying the generalization ability of the proposed approach under cross-database testing scenarios; and (iii) compiling the LFW+ database\footnotemark[2] with face images in the wild (LFW), and heterogeneous demographic attributes (age, gender, and race) via crowdsourcing.\footnotetext[2]{We plan to place the LFW+ dataset in the public domain.}


\begin{table*}[!ht]
\begin{center}
\caption{A summary of published methods on multi-attribute estimation from a face image.}
\label{tab:MultiAttributeMethods}
\vspace{-1mm}
\scriptsize
\begin{tabular}{llll}
\toprule
\textbf{Publication} & \tabincell{c}{\textbf{Approach}\\ (feature and prediction model)} & \tabincell{c}{\textbf{Face database} \\ \#images (training; testing)} & \tabincell{c}{\textbf{Accuracy}} \\
\hline
\tabincell{l}{Cottrell and Metcalfe \\ \cite{CottrellNIPS90}} & \tabincell{l}{Autoencoder; \\One backpropagation network per attribute} & \tabincell{l}{Private dataset \\ ($160$, $40$)} & \tabincell{l}{\textbf{Private dataset} \\ Emotion: $<50\%$ (Avg. of eight classes) \\ Gender: 100\% (on training set)}\\

\hline
\tabincell{l}{Kumar \etal \\ \cite{KumarECCV08}}& \tabincell{l}{Grayscale and color pixel values, \\edge magnitude, and gradient direction; \\One SVM classifier per attribute} & \tabincell{l}{CelebA (public) \\ ($180K$, $20K$)\\ LFW$^1$ (public)\\(n/a; 13,143)} & \tabincell{l}{\textbf{CelebA}: 81\% (Avg. of 40 attrs.) \cite{LiuICCV15} \\ \textbf{LFW}$^1$ Gender: $92.7\%$; Race: $90.3\%$}  \\

\hline



\tabincell{l}{Guo and Mu  \cite{GuoIVC14}} & \tabincell{l}{Biologically-inspired features (BIFs);\\ multi-label regression with CCA and PLS} & \tabincell{l}{MORPH II (public) \\ (10,530, 44,602)} & \tabincell{l}{\textbf{MORPH II} \\ Age: $70.0\%$ CS(5)$^2$, 3.92 yrs. MAE \\ Gender: 98.5\%, Race: 99.0\% (Black vs. White)} \\
\hline

\tabincell{l}{Yi \etal \cite{YiACCV14}} & \tabincell{l}{Concatenated features by multi-scale CNN \\ (3-layer network); \\multi-label loss} & \tabincell{l}{MORPH II (public) \\ (10,530, 44,602)} & \tabincell{l}{\textbf{MORPH II} \\ Age: 3.63 yrs. MAE \\ Gender: 98.0\%, Race: 99.1\% (Black vs. White)} \\
\hline

Eidinger \etal \cite{EidingerTIFS14} &  \tabincell{l}{LBP and four-patch LBP;\\ One SVM classifier per attribute} & \tabincell{l}{\emph{Images of Groups} (public)\\($3,500$; $1,050$)\\ Adience (public)\\(13,000; 3,300)} &  \tabincell{l}{\textbf{Images of Groups}\\Age group: 66.6\%, Gender: 88.6\% \\ \textbf{Adience} \\ Age group: 45.1\%, Gender: 76.1\%} \\
\hline

Han \etal \cite{HanTPAMI15} & \tabincell{l}{BIFs with feature selection;\\ One SVM classifier per attribute} & \tabincell{l}{MORPH II (public) \\(20,569; 78,207)\\PCSO (private) \\ (81,533; 100,012) \\LFW Frontal (public)\\ (4211, 4211)} & \tabincell{l}{\textbf{MORPH II} Age: $77.4\%$ CS(5)$^2$, $3.6$ yrs. MAE \\ Gender: 97.6\%, Race: 99.1\% (Black vs. White) \\ \textbf{PCSO} Age: $72.6\%$ CS(5)$^2$, $4.1$ yrs. MAE \\ Gender: 97.1\%, Race: 98.7\% (Black vs. White) \\ \textbf{LFW Frontal} Age: $42.5\%$ CS(5)$^2$, $7.8$ yrs. MAE \\ Gender: 94\%, Race: 90\% (White vs. Other)}\\
\hline

\tabincell{l}{Levi and Hassner \\ \cite{LeviCVPRW15}} & \tabincell{l}{CNN with 3 Conv. layers and 2 FC layers;\\ One CNN classifier per attribute} & \tabincell{l}{Adience (public) \\(15,590; 3,897)$^3$} & \tabincell{l}{\textbf{Adience} \\ Age group: 50.7\%, Gender: 86.8\%}\\ 
\hline

Liu \etal \cite{LiuICCV15} & \tabincell{l}{Multi-patch features by a cascade of \\  LNets (5 Conv. layers) and ANet \\(4 Conv. layers); \\One SVM classifier per attribute} &\tabincell{l}{CelebA (public) \\ ($180K$, $20K$) \\ LFWA (public)\\ ($6,263$; $6,970$)} & \tabincell{l}{\textbf{CelebA}: 87\% (Avg. of 40 attributes) \\ \textbf{LFWA}: 84\% (Avg. of 40 attributes)}  \\
\hline

Huang \etal \cite{HuangCVPR16} & \tabincell{l}{CNN features by DeepID2 with large \\margin local embedding; kNN classifier} & \tabincell{l}{CelebA (public) \\ ($180K$, $20K$)} & \tabincell{l}{\textbf{CelebA}: 84\%$^4$ (Avg. of 40 attributes)} \\
\hline

U\v{r}i\v{c}\'{a}\v{r} \etal \cite{UricarCVPRW16} & \tabincell{l}{CNN features by VGG-16 \cite{SimonyanArXiv2014};\\One SVM classifier per attribute} & \tabincell{l}{ChaLearn LAP 2016 (public) \\(4,113; 1500 (validation set))} & \tabincell{l}{\textbf{ChaLearn LAP2016 (validation set)}\\ Age: 0.24 $\epsilon$-error, Gender: 89.2\%, \\Smile: 79.03\%}\\
\hline

Ehrlich \etal \cite{EhrlichCVPRW16} & \tabincell{l}{Multi-task Restricted Boltzmann \\ Machines with PCA and keypoint \\ features; \\ Multi-task classifier} & \tabincell{l}{CelebA (public) \\ ($180K$, $20K$) \\ ChaLearn FotW \\ (6,171; 3,087)} & \tabincell{l}{\textbf{CelebA}: 87\% (Avg. of 40 attributes)\\\textbf{FotW}: Smile and gender: 76.3\% (Avg.)} \\
\hline

\tabincell{l}{Hand and Chellappa \\ \cite{HandArXiv2016}} & \tabincell{l}{Multi-task CNN features (3 Conv. \\layers and 2 FC layers); \\ Joint regression of multiple binary \\attributes} & \tabincell{l}{CelebA (public) \\ ($180K$, $20K$) \\ LFWA (public)\\ ($6,263$; $6,970$)} & \tabincell{l}{\textbf{CelebA} 91\% (Avg. of 40 attributes)\\\textbf{LFWA} 86\% (Avg. of 40 attributes)} \\
\hline

Zhong \etal \cite{ZhongICB16} & \tabincell{l}{Off-the-shelf CNN features by \\ FaceNet and VGG-16 \cite{SimonyanArXiv2014}\\ One SVM classifier per attribute} & \tabincell{l}{CelebA (public) \\ ($180K$, $20K$)\\ LFWA (public) \\ ($6,263$; $6,970$)} &  \tabincell{l}{\textbf{CelebA} 86.6\% (Avg. of 40 attributes) \\ \textbf{LFWA} 84.7\% (Avg. of 40 attributes)}  \\
\hline

Proposed method & \tabincell{l}{Deep multi-task feature learning (DMTL)\\ with shared feature learning (modified \\AlexNet) and category-specific feature \\learning (2 FC layers) \\ Joint estimation of multiple heterogeneous \\attributes} & \tabincell{l}{MORPH II (public) \\ ($62,566$; $15,641$)$^3$ \\  LFW+ (created by authors) \\($12,559$; $3,140$)$^3$ \\ CelebA (public) \\ ($180K$, $20K$) \\ LFWA (public) \\ (6,263; 6,970) \\ LAPAge2015 (public)\\(2,476; 1,136)\\ChaLearn FotW (public) \\ (6,171; 3,087)} & \tabincell{l}{\textbf{MORPH II} (w/o pre-training on IMDB-WIKI)\\Age: $85.3\%$ CS(5)$^2$, 3.0 yrs. MAE; \\Gender: 98.0\%, Race: 96.6\% (Black, White, Other)\\ \iffalse \textbf{MORPH II} (with fine-tuning on IMDB-WIKI)\\ Age: $??\%$ CS(5)$^2$, ?? yrs. MAE; \\Gender: ??\%, Race: ??\% (Black, White, Other) \\ \fi \textbf{LFW+} Age: $75.0\%$ CS(5)$^2$, $4.5$ yrs MAE; \\ Gender: $96.7\%$; Race: $94.9\%$ \\ \textbf{CelebA} 92.1\% (Avg. of 40 attributes); \\ \textbf{LFWA} 86\% (Avg. of 40 attributes) \\ \textbf{CLAP2015} (w/o pre-training on IMDB-WIKI)\\ Age: 5.2 yr. MAE, $\epsilon$-error: 0.449 \\ \textbf{FotW} Accessory: 94.0\% (Avg. of 7 attributes); \\Smile and gender: 86.1\% (Avg.)}  \\


\bottomrule
\end{tabular}
\end{center}
\vspace{-1mm}
\footnotesize
$^1$The ground-truth age, gender, and race information of the LFW face images was not provided in \cite{KumarECCV08}; the accuracies reported for \cite{KumarECCV08} are from \cite{LiuICCV15}.
$^2$CS(5) denotes the age estimation accuracy @ 5-year absolute error.
$^3$The numbers of training and testing images reported here are the average in one-fold test.
$^4$A different metric is used: an average of true positive rate and true negative rate.

\vspace{-6mm}
\end{table*}

Some of the preliminary work is described in \cite{HanTR14,WangFG17}. %
Essential improvements in this work include: (i) extensions in category-specific feature learning for handling attribute heterogeneities in terms of data type and scale, and semantic meaning; (ii) additional technical and implementation details; and (iii) extensive evaluations using 6 different attribute databases, and comparisons with additional state of the art.

The remainder of this paper is structured as follows.
In Section~\ref{Sec.RelatedWork}, we briefly review related literature.
In Section~\ref{Sec.ProposedApproach}, we detail the proposed heterogeneous face attribute estimation approach.
In Section~\ref{Sec.Experiment}, we introduce the LFW+ database which contains faces in the wild, and heterogeneous attributes of age, gender, and race obtained via crowdsourcing, and provide the experimental results and analysis.
Finally, we conclude this work in Section~\ref{Sec.Summary}.

\section{Related Work}
\label{Sec.RelatedWork}

\subsection{Multi-attribute Estimation From Face}
\label{Sec.JointAttributeEstMethod}

While there are a number of studies on face attribute estimation in the literature, many of them focus on estimating a single attribute, \eg age, expression, etc.
The age estimation error with mean absolute error (MAE) metric has been reduced by a large margin from $8.8$ years \cite{GengPAMI07} to $2.68$ years \cite{RotheIJCV16} on the MORPH II database \cite{RicanekFG06}.
Facial expression recognition accuracy has been substantially improved from less than $80\%$ to over than $93\%$ on the {Cohn-Kanade} database \cite{KanadeFG00,ZengTPAMI08}.
Due to limited space, we refer interested readers to reviews of the prior work on single facial attribute estimation in \cite{MakinenPRL08,ZengTPAMI08,FuPAMI10,GengPAMI13,WangWACV15,HanTPAMI15,PanisIETB16,RotheIJCV16}.
In the following, we briefly review the most recent literature on joint estimation of multiple face attributes, covering feature representation, prediction models, databases, and performance (see Table~\ref{tab:MultiAttributeMethods}).

Attempts to design computational models based on psychological studies on multi-attribute estimation from a face image started in the 1990s \cite{CottrellNIPS90}.
Since then, a number of approaches have been reported in the literature, but the early work utilized hand-crafted features for attribute estimation.
In \cite{KumarECCV08}, edge magnitude and gradient features were extracted from various face regions; the same features were used to learn a separate SVM classifier for each face attribute.
Multi-label regressions using canonical correlation analysis (CCA) and partial least squares (PLS) based on BIF features were used in \cite{GuoIVC14} for joint estimation of three face attributes (age, gender, and race); the joint estimation resulted in a better performance than separate models for age, gender, and race.
In \cite{EidingerTIFS14}, per attribute dropout-SVM classifiers were trained using LBP features for estimating age and gender, respectively.
BIF features with three separate SVM classifiers were used for age, gender, and race estimation in \cite{HanTPAMI15}, but unlike \cite{GuoIVC14}, feature selection was applied to BIF features to find demographic informative features for each attribute.

Except for \cite{CottrellNIPS90} which used autoencoder for feature learning, all the above approaches utilized hand-crafted features.
Recently, the biologically inspired deep learning network has resulted in significant advances in many computer vision tasks \cite{LeCunNature15}, including face attribute prediction, due to their ability to learn compact and discriminative features \cite{YiACCV14,LiICML14,LiuICCV15}. 
In \cite{YiACCV14}, CNN features extracted from multi-scale patches were concatenated together, and used for joint estimation of three face attributes (age, gender, and race).
A CNN with three convolutional layers and two FC layers was proposed in \cite{LeviCVPRW15}, and per attribute CNNs were trained to handle age and gender estimation, respectively.
In \cite{LiuICCV15}, a cascaded network of face localization (LNet) and attribute prediction (ANet) was used for face localization and feature extraction for individual SVM classifiers.
Additionally, two face attribute databases (CelebA and LFWA) were presented in \cite{LiuICCV15} along with face image labels.
Per attribute SVM classifiers were also used in \cite{UricarCVPRW16} for the estimation of age, gender, and smile, but the features were learned using the VGG-16 network \cite{SimonyanArXiv2014}.
In \cite{ZhongICB16}, a similar idea of per attribute SVM classifiers using FaceNet and VGG-16 features was applied for estimating $40$ face attributes on the CelebA and LFWA databases \cite{LiuICCV15}.
In \cite{HuangCVPR16}, a large margin local embedding kNN (LMLE-kNN) approach was proposed to deal with large-scale imbalanced attribute classification tasks.
With PCA appearance features and keypoint features, Multi-task Restricted Boltzmann Machine (RBM) was adopted in \cite{EhrlichCVPRW16} for estimating $40$ face attributes on the CelebA database, and gender and smile classifications on the FotW database \cite{EscaleraCVPRW16}.


\subsection{Multi-Task Learning in Deep Networks}
\label{Sec.MTL}

As summarized in Table~\ref{tab:MultiAttributeMethods}, approaches using hand-crafted and deep learning features can be grouped into two categories: (i) single-task learning (STL) of per attribute classifier \cite{CottrellNIPS90,KumarECCV08,EidingerTIFS14,HanTPAMI15,LeviCVPRW15,LiuICCV15,UricarCVPRW16,ZhongICB16}; and (ii) multi-task learning (MTL) of a joint attribute classifier \cite{EhrlichCVPRW16}.
Compared with STL based methods, where each attribute is estimated separately, ignoring any correlations between the tasks, MTL based methods learn multiple models for multi-attribute estimation using a shared representation \cite{CaruanaML97}.
Deep models are well suited for MTL; therefore, a number of approaches seek to combine MTL with deep learning.
Besides the MTL networks for face attribute estimation \cite{YiACCV14,EhrlichCVPRW16,HandArXiv2016}, MTL networks have been proposed for human pose estimation \cite{LiIJCV2015}, human attribute prediction \cite{AbdulnabiTMM15}, face alignment \cite{ZhangTPAMI16,SagonasICV16}, etc.

The proposed approach falls under the MTL approach with CNNs, but with several differences compared with existing methods \cite{LiuICCV15,ZhangTPAMI16,LiIJCV2015,AbdulnabiTMM15,EhrlichCVPRW16,HandArXiv2016}.
\begin{itemize}
  \item Unlike existing methods that have focused on face alignment, human pose estimation, and human attribute estimation \cite{ZhangTPAMI16,LiIJCV2015,AbdulnabiTMM15}, the proposed approach focuses on joint estimation of multiple attributes from a face image.
  \item Unlike the MTL in \cite{ZhangTPAMI16} which utilizes the auxiliary tasks to assist in the main task, we aim to boost the estimation accuracies of all the face attributes through utilizing attribute correlations and handling attribute heterogeneities;
  \item Unlike the methods in \cite{LiuICCV15,HuangCVPR16} which utilized a two-step pipeline of CNN features followed by attribute classifiers, the proposed DMTL is an end-to-end learning approach;
  \item The proposed approach considers a number of practical scenarios for heterogeneous attribute estimation, single attribute estimation, and cross-database testing.
\end{itemize}


\section{Proposed Approach}
\label{Sec.ProposedApproach}

\subsection{Deep Multi-task Learning}
\label{Sec.DMTL}

\begin{figure}[!t]
\includegraphics[width=0.48\textwidth,height=.43\textwidth]{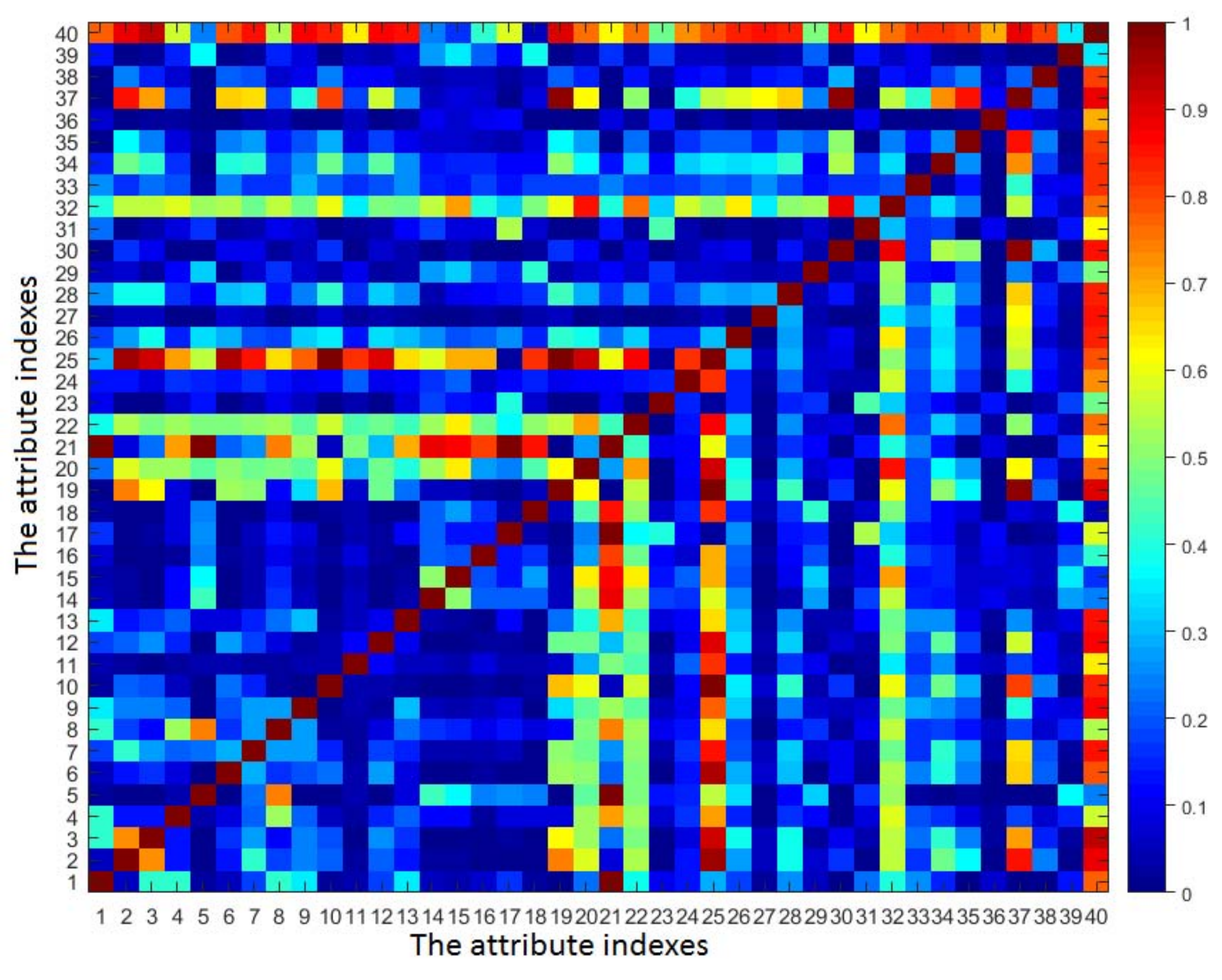}
\vspace{-4mm}
\caption[Co-occurrence]{Pair-wise co-occurrence matrix of the $40$ face attributes (see Table \ref{tab:CelebAttrInx}) provided with the CelebA database\footnotemark[3]. Examples of attributes with a strong positive correlation include: \#1 (5 O'Clock Shadow) and attribute \#21 (Male), and attribute \#19 (Heavy Makeup) and \#37 (Wear Lipstick).}
\label{Fig.Cooccurrence}
\vspace{-6mm}
\end{figure}

Our aim is to simultaneously estimate a large number of face attributes via a joint estimation model.
While a large number of face attributes pose challenges to the feature learning efficiency, they also provide opportunities for leveraging the attribute inter-correlations to obtain informative and robust feature representation.
For example, as shown in Fig. \ref{Fig.Cooccurrence}, a number of attributes in the CelebA database \cite{LiuICCV15} have strong pair-wise correlations (elements with red color).
MTL methods are naturally suited for this joint estimation problem.
However, presence of appearance variations in facial images and the heterogeneity of individual attributes, the mapping from the face image space to the attribute space is typically nonlinear.
Therefore, the joint attribute estimation model should also be able to capture the complex and compositional nonlinear transformation between its input and output.
CNN model is an effective approach for handling both MTL and such a nonlinear transformation learning.
A good overview of MTL in neural network can be found in \cite{CaruanaML97}.
Following the success of MTL in neural networks \cite{LiIJCV2015,AbdulnabiTMM15,ZhangTPAMI16}, we choose to use Deep Multi-task Learning (DMTL) for estimating multiple attributes from a single face image.

\footnotetext[3]{\url{http://mmlab.ie.cuhk.edu.hk/projects/CelebA.html}}

We assume a training dataset with $N$ face images, each with $M$ attributes. The dataset is denoted as $\mathbf{D} = \{\mathbf{X}, \mathbf{Y}\}$, where $\mathbf{X} = \{X_i\}_{i=1}^{N}$, and $\mathbf{Y} = \left\{\{y_{i}^{j}\}_{j=1}^{M}\right\}_{i=1}^{N}$.
A traditional DMTL model for joint attribute estimation can be formulated by minimizing the regularization error function
\begin{equation}
\mathop{\arg\min}_{\{W^j\}_{j=1}^{M}} \sum_{j=1}^{M}  \sum_{i=1}^{N}  \mathcal{L}  \big(y_i^{j}, \mathcal{F}(X_i, W^j)\big) + \gamma\Phi(W^j),
\label{Eq:TradMTL}
\end{equation}
where $\mathcal{F}(\cdot,\cdot)$ is an attribute prediction function of the input $X_i$ and weight vector $W^j$;
$\mathcal{L}(\cdot, \cdot)$ is a prescribed loss function (\eg empirical error) between estimated values by $\mathcal{F}$ and the corresponding ground-truth values $y_i^{j}$;
$\Phi(\cdot)$ is a regularization term which penalizes the complexity of weights, and $\gamma$ is a regularization parameter ($\gamma >0$).

\begin{figure}[!t]
\includegraphics[width=0.48\textwidth,height=.3\textwidth]{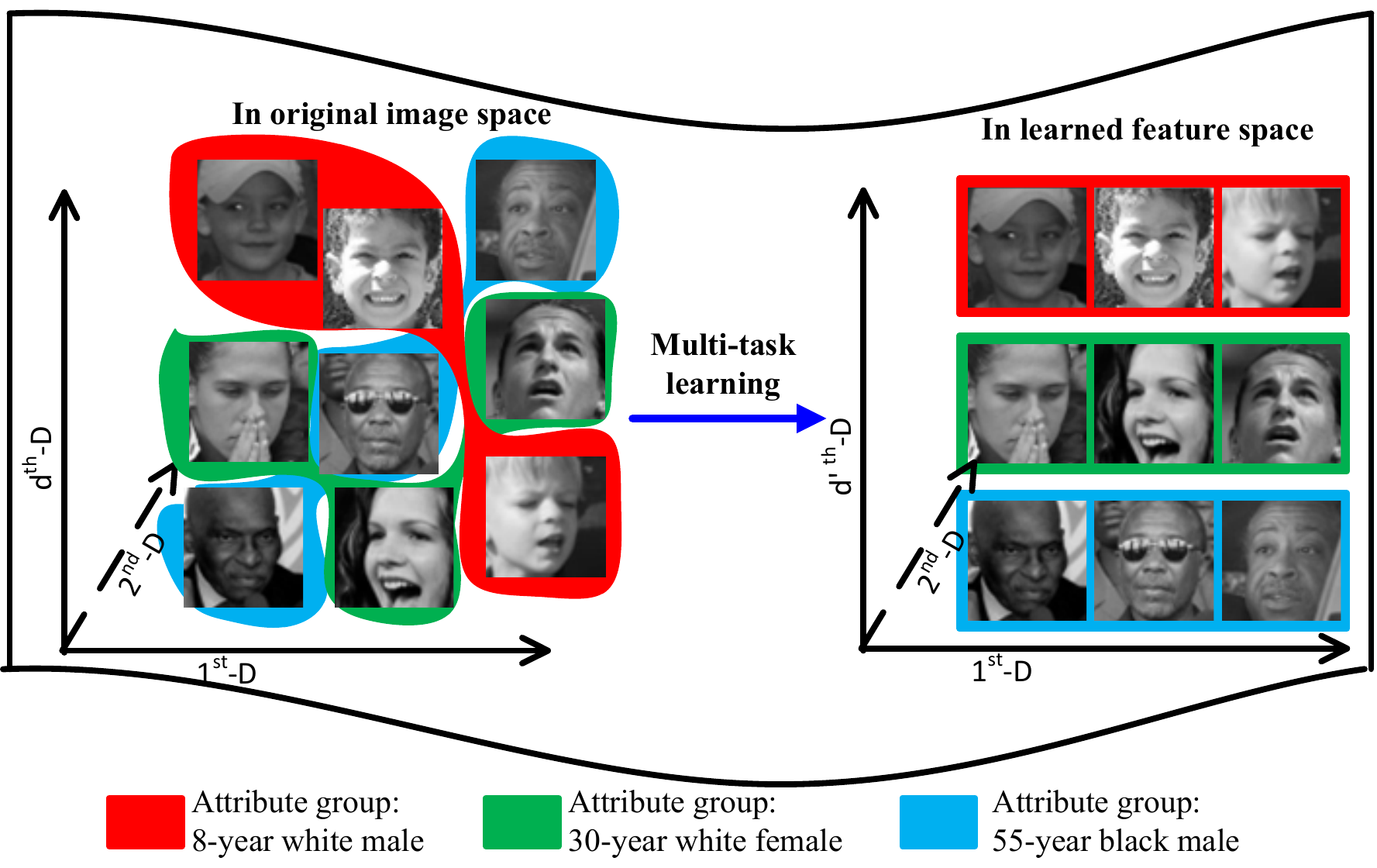}
\vspace{-2mm}
\caption[EffectofMTL]{The benefit of using MTL is that individual attribute groups which are not well separable from each other in the original image space could become separable in the feature space learned by MTL, leading to improved multi-attribute estimation accuracy.}
\label{Fig.EffectofMTL}
\vspace{-6mm}
\end{figure}

Given the objective function in (\ref{Eq:TradMTL}), a straightforward approach is to learn multiple CNNs in parallel, one per attribute.
Such an approach is not optimal because individual face attribute estimation tasks may share some common features.
This is supported by the fact that off-the-shelf CNN features learned for face recognition were directly used for face attribute estimation \cite{ZhongICB16}.
However, the formulation in (\ref{Eq:TradMTL}) does not explicitly enforce a large portion of feature sharing during MTL.
To this end, we reformulate the DMTL for multi-attribute estimation as
\begin{equation}
\begin{aligned}
 \mathop{\arg\min}_{W_c,\{W^j\}_{j=1}^{M}} \sum_{j=1}^{M} \sum_{i=1}^{N}  \mathcal{L}  \big(y_i^{j}, \mathcal{F}(X_i, W^{j} \circ W_c)\big) \\
+ \gamma_1\Phi(W_c) + \gamma_2\Phi(W^j)
 \end{aligned},
 \label{Eq:FaceDMTL}
\end{equation}
where $W_c$ controls feature sharing among the face attributes, and $W^j$ controls update of the shared features w.r.t. each face attribute.
Specifically, as shown in Fig. \ref{Fig.DMTL}, a face image is first projected to a high-level representation through a shared deep network ($W_c$) consisting of a cascade of complex non-linear mappings, and then refined by shallow subnetworks ($\{W^j\}_{j=1}^{M}$) towards individual attribute estimation tasks.
The formulation in (\ref{Eq:FaceDMTL}) makes it possible to explore the attribute correlations and learn a compact representation shared by various attributes.
Figure \ref{Fig.EffectofMTL} explains the benefit of jointly estimating multiple face attributes via MTL.

\subsection{Heterogeneous Face Attribute Estimation}
\label{Sec.HeteEst}

Although the above formulation of DMTL utilizes the attribute correlations in feature learning, the attribute heterogeneity still needs to be considered.
Heterogeneity of individual face attribute is ever present, but has not received sufficient attention.
The reasons are two-fold: (i) many of the public-domain face databases are labeled with a single attribute, the requirement of designing corresponding models becomes no longer urgent.
(ii) many of the published methods choose to learn a separate model for each face attribute; model learning for individual attributes does not face the attribute heterogeneity problem.

We treat each of the heterogeneous attribute categories separately, but attributes within each category are expected to share feature learning and classification model to a larger extent.
To accomplish this, the objective function in (\ref{Eq:FaceDMTL}) is rewritten as
\begin{equation}
\begin{aligned}
 \mathop{\arg\min}_{W_c, \{W^j\}_{j=1}^{M}} \sum_{g=1}^{G}\sum_{j=1}^{M^g} \sum_{i=1}^{N} \lambda^g \mathcal{L}^{g}  \big(y_i^{j}, \mathcal{F}(X_i, W^{g} \circ W_c) \big) \\
+ \gamma_1\Phi(W_c) + \gamma_2\Phi(W^g)
 \end{aligned},
 \label{Eq:ProposedDMTL}
\end{equation}
where $G$ is the number of heterogeneous attribute categories, and $M^g$ is the number of attributes within each attribute category;
$\lambda^g$ balances the importance of each attribute category ($\lambda^g =1$ by default);
$W^g$ refines the shared features w.r.t. each of the heterogeneous attribute categories.
$\mathcal{L}^g(\cdot, \cdot)$ is a prescribed loss function for each of the heterogeneous attribute categories, given the estimated values by $\mathcal{F}$ and the corresponding ground-truth $y_i^{j}$.

Grouping a large number of attributes into a few heterogeneous categories depends on prior knowledge.
Here, we consider face attribute heterogeneities in terms of \emph{data type and scale} (\ie ordinal vs. nominal) \cite{Jain88} and \emph{semantic meaning} (\ie holistic vs. local) \cite{SamangooeiBTAS08}, and explain our category-specific modeling for these heterogeneous attribute categories.

\textbf{Nominal vs. ordinal attributes}. Nominal attributes have two or more classes (values), but there is no intrinsic ordering among the categories \cite{Jain88}.
For example, race is a nominal attribute having multiple classes, such as Black, White, Asian, etc., and there is no intrinsic ordering of these values (classes).
We handle nominal attributes in a classification scheme, and choose to use the cross-entropy loss \cite{deBoerAOR05}

\begin{equation}
\begin{split}
\mathcal{L}^{g_{\mathbb{N}}} = -\sum_{j=1}^{M^{\mathbb{N}}} \sum_{i=1}^{N} \sum_{k=1}^{C^j} \mathbf{1}(y_i^{j}, \hat{y}_i^{j,k}) \log p(\hat{y}_i^{j,k} )
\end{split},
\label{Eq:CELoss}
\end{equation}
where
\begin{equation}
\begin{split}
p(\hat{y}_i^{j,k}) = \frac{e^{\hat{y}_i^{j,k}}}{\sum_{k=1}^{C^j} e^{\hat{y}_i^{j,k}}}
\end{split}
\label{Eq:Softmax}
\end{equation}
is the softmax function, $\hat{y}_i^{j,k}$ is the $k$-th element ($C^j$ elements in total) of the prediction by $\mathcal{F}(X_i, W^{g_{\mathbb{N}}} \circ W_c)$ for the estimation of the $j$-th nominal attribute; $y_i^{j}$ is the ground-truth attribute;
and $\mathbf{1}(a,b)$ outputs $1$ when $a = b$, and $0$ otherwise.


The difference between ordinal attribute and nominal attribute is that ordinal attribute has a clear ordering of its variables.
For example, age of a person, typically ranging from $0$ to $100$, is an ordinal attribute..
Actually, age is not only ordinal but also interval \cite{Jain88}.
We handle ordinal attributes in a regression scheme, and choose to use the Euclidean loss
\begin{equation}
\begin{split}
\mathcal{L}^{g_{\mathbb{O}}} = \sum_{j=1}^{M^{\mathbb{O}}} \sum_{i=1}^{N} \|y_i^{j} - \hat{y}_i^{j}\|_2^2
\end{split},
\label{Eq:L2Loss}
\end{equation}
where $\hat{y}_i^{j,k}$ is the prediction by $\mathcal{F}(X_i, W^{g_{\mathbb{O}}} \circ W_c)$.

\textbf{Holistic vs. local attributes}. While attributes such as age, gender, and race describe the characteristics of the whole face, attributes such as pointy nose and big lips, mainly describe the characteristics of local facial components.
Therefore, the optimal features used for estimating holistic and local attributes could be different.
Such attribute heterogeneities in semantic meaning can also be modeled by the proposed DMTL, \eg using multiple holistic attribute subnetworks and multiple per-component attribute subnetworks.
Both holistic and local attributes could further consist of nominal and ordinal categories.
So, a joint consideration of nominal vs. ordinal and holistic vs. local heterogeneities leads to four types of subnetworks: holistic-nominal, holistic-ordinal, local-nominal, and local-ordinal.
The choice of the loss function for each type of subnetwork is still determined by whether the subnetwork is nominal or ordinal.

The proposed category-specific modeling differs the proposed approach from \cite{HandArXiv2016}, which manually classifies the binary attributes into $9$ groups based on the attribute location (\eg eyes, nose, and mouth), but does not consider the heterogeneity in terms of data type and scale.
In addition, each of the $9$ attribute groups in \cite{HandArXiv2016} was handled equally via regression.

\subsection{Implementation Details}
\label{Sec.ImpDetails}

As shown in Fig. \ref{Fig.DMTL}, the proposed DMTL network mainly consists of a deep network for shared feature learning, and variable number of shallow subnetworks for category-specific feature learning.

\emph{Network structure.}
For the shared feature learning network, we use a modified AlexNet network (5 convolutional (Conv.) layers, 5 pooling layers, 2 fully connected (FC) layers \cite{KrizhevskyNIPS12}) with a batch normalization (BN) layer inserted after each of the Conv. layers.
Each of the category-specific feature learning networks contains two FC layers, and is connected to the last FC layer of the shared network.

\begin{figure}[!t]
\includegraphics[width=0.49\textwidth,height=.18\textwidth]{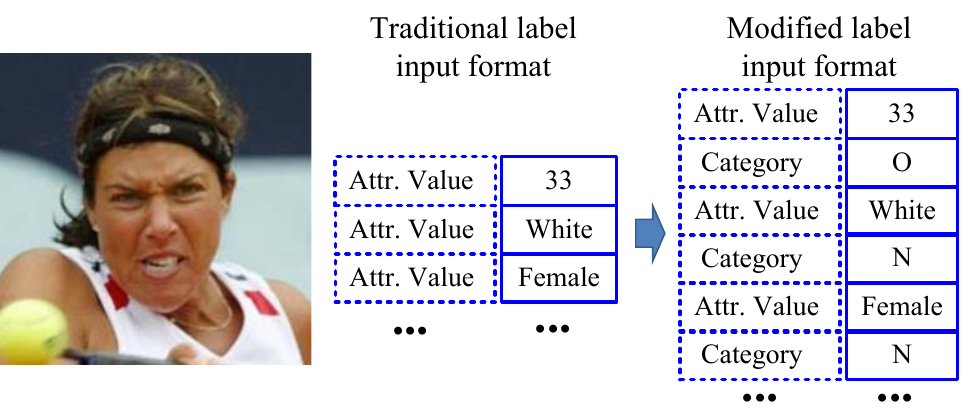}
\vspace{-8mm}
\caption[LabelFormat]{Revised network input for the label information, with each attribute taking two fields: one for the attribute value and the other for attribute category. Here, `N' and `O' represent the nominal and ordinal attributes, respectively.}
\label{Fig.LabelFormat}
\vspace{-6mm}
\end{figure}

\emph{Network input.} Since the proposed DMTL network is designed to handle heterogeneous attribute categories, we revise the network input format,
and use two fields to represent each attribute label, \ie $y_i^j = (val, cat)$, where $val$ and $cat$ denote the attribute value and category, respectively (see Fig. \ref{Fig.LabelFormat}).
After we introduced an attribute category field, the order of the input attributes no longer matters; the corresponding attribute values used for computing individual losses (cross-entropy and Euclidean) can be easily determined based on the attribute category fields, e.g.,  ${cat = N}$ for choosing the nominal attribute values, and ${cat = O}$ for choosing the ordinal attribute values.
This is an advantage of the proposed approach over existing methods like \cite{LiIJCV2015,HandArXiv2016}.

\emph{Network training.} We perform stochastic gradient descent (SGD) \cite{KrizhevskyNIPS12} with weight decay  \cite{KroghNIPS91} to jointly optimize the weights of both the shared network and the category-specific subnetworks in an end-to-end way.
Specifically, given two types of loss functions (for ordinal and nominal attributes), the derivatives used for updating $W^{g_{\mathbb{N}}}$ and $W^{g_{\mathbb{O}}}$ can be calculated as
\begin{equation}
\begin{split}
\frac{\partial \mathcal{L}^{g_{\mathbb{N}}}}{\partial W^{g_{\mathbb{N}}}} = \big( y_i^{j} - p(\hat{y}_i^{j,k}) \big) {X_i}^{\mathrm{T}}
\end{split},
\label{Eq:PartialCE}
\end{equation}
and
\begin{equation}
\begin{split}
\frac{\partial \mathcal{L}^{g_{\mathbb{O}}}}{\partial W^{g_{\mathbb{O}}}} = \big( y_i^{j} - (W^{g_{\mathbb{O}}})^\mathrm{T} X_i \big) {X_i^{\mathrm{T}}}
\end{split}.
\label{Eq:PartialL2}
\end{equation}
The sum of (\ref{Eq:PartialCE}) and (\ref{Eq:PartialL2}) is used for updating $W_c$.
Finally, the network weights are updated as
\begin{equation}
\begin{split}
&\Delta W^{g_{\mathbb{N}}} = \eta \frac{\partial \mathcal{L}^{g_{\mathbb{N}}}}{\partial W^{g_{\mathbb{N}}}},\\
&\Delta W^{g_{\mathbb{O}}} = \eta \frac{\partial \mathcal{L}^{g_{\mathbb{O}}}}{\partial W^{g_{\mathbb{O}}}},\\
&\Delta W^{c} = \eta \big( \frac{\partial \mathcal{L}^{g_{\mathbb{N}}}}{\partial W^{g_{\mathbb{N}}}} + \frac{\partial \mathcal{L}^{g_{\mathbb{O}}}}{\partial W^{g_{\mathbb{O}}}} \big), \\
\end{split}
\label{Eq:PartialL2}
\end{equation}
where $\eta$ is the learning rate.
Random initialization is used for all the weights in network pre-training.


\section{Experimental Results}
\label{Sec.Experiment}

\subsection{Databases}
\label{Sec.Databases}

As summarized in Section \ref{Sec.RelatedWork}, the widely used public-domain face database for attribute estimation include: MORPH II \cite{RicanekFG06}, CelebA \cite{LiuICCV15}, LFWA \cite{LiuICCV15}, and ChaLearn LAP\cite{EscaleraICCVW15} and FotW \cite{EscaleraCVPRW16}.
Besides these databases, we also constructed the LFW+ database (LFW augmented by $2,466$ images of children) with three heterogeneous attributes, labeled for each face image via the Amazon Mechanical Turk (MTurk) crowdsourcing\footnotemark[4]\footnotetext[4]{\url{https://www.mturk.com}}.

\begin{table}[!t]
\begin{center}
\caption{Summary of the $40$ face attributes provided with the CelebA database \cite{LiuICCV15}.}
\label{tab:CelebAttrInx}
\vspace{-1mm}
\footnotesize
\begin{tabular}{c|l|c|l}
\toprule
\textbf{Attr. Idx.}&    \textbf{Attr. Def.}&	   \textbf{Attr. Idx.}&    \textbf{Attr. Def.}	   \\
\midrule
1& 5 O'ClockShadow & 21 & Male  \\
2& ArchedEyebrows  & 22 & MouthSlightlyOpen \\
3& BushyEyebrows   & 23 & Mustache\\
4& Attractive      & 24 & NarrowEyes\\
5& BagsUnderEyes   & 25 & NoBeard\\
6& Bald            & 26 & OvalFace\\
7& Bangs           & 27 & PaleSkin\\
8& BlackHair       & 28 & PointyNose\\
9& BlondHair       & 29 & RecedingHairline\\
10& BrownHair      & 30 & RosyCheeks\\
11& GrayHair       & 31 & Sideburns  \\
12& BigLips        & 32 & Smiling \\
13& BigNose        & 33 & StraightHair\\
14& Blurry         & 34 & WavyHair\\
15& Chubby         & 35 & WearEarrings\\
16& DoubleChin     & 36 & WearHat\\
17& Eyeglasses     & 37 & WearLipstick\\
18& Goatee         & 38 & WearNecklace\\
19& HeavyMakeup    & 39 & WearNecktie\\
20& HighCheekbones & 40& Young\\
\bottomrule
\end{tabular}
\end{center}
\vspace{-6mm}
\end{table}

\textbf{MOROH II.} MORPH is a large database of mugshot images, each with associated metadata containing three heterogeneous attributes: age (ordinal), gender (nominal), and race (nominal).
We investigate all the three attribute estimation tasks on MORPH Album2 (MORPH II) containing about $78K$ images of more than $20K$ subjects.
Results on MORPH II are reported with a five-fold, subject-exclusive cross-validation protocol \cite{HanTPAMI15,RotheIJCV16}.

\textbf{CelebA.} CelebA is a large-scale face attribute database \cite{LiuICCV15} with more than $200K$ celebrity images of more than $10K$ identities, each with $40$ attribute annotations (see Table \ref{tab:CelebAttrInx}).
The images in this dataset contain large variations in pose, expression, race, background, etc., making it challenging for face attribute estimation.
Additionally, since there are $40$ attribute annotations, the CelebA database poses challenges to joint attribute estimation algorithms in terms of feature learning efficiency.
Results on CelebA are reported following the protocol provided in \cite{LiuICCV15}.

\textbf{LFWA.} LFWA is another unconstrained face attributes database \cite{LiuICCV15} with face images from the LFW database ($13,233$ images of $5,749$ subjects) \cite{HuangTR07}, and the same $40$ attribute annotations as in the CelebA database.
Results on LFWA are reported following the protocol provided in \cite{LiuICCV15}.

\textbf{ChaLearn LAP and FotW.} The ChaLearn challenge series, started in $2011$, has been very successful in promoting advances in visual or multi-modal analysis of people \cite{EscaleraArXiv2017}.
LAPAge2015 is an unconstrained face database for apparent age estimation released at ICCV 2015.\footnotemark[5]\footnotetext[5]{\url{http://gesture.chalearn.org/2015-looking-at-people-iccv-challenge}}
This database contains $4,699$ face images, each with an average age of the estimates by at least $10$ different users.
The database was split into $2,476$ images for training, $1,136$ images for validation, and $1,087$ images for testing \cite{EscaleraICCVW15}.
Since the age information for the testing set was not available, we follow the protocol in \cite{RotheIJCV16}, and report the results on the validation set.
The FotW database was created by collecting publicly-available images from the Internet, which contains two datasets, one for accessary classification, and the other for gender and smile classification.
The FotW accessary dataset contains $5,651$, $2,826$, and $4,086$ face images for training, validation, and testing, respectively; each is annotated with seven binary accessory attributes (see Table \ref{tab:FotWAccuracy} (a)).
The FotW gender and smile dataset is composed of $6,171$, $3,086$, and $8,505$ face images for training, validation, and testing, respectively; each is annotated with ternary gender (male, female, and not sure) and binary smile attributes.
We following the same testing protocols to report the results on FotW.

\textbf{LFW+.} We extended the LFW database \cite{HuangTR07} to study the joint attribute estimation (age, gender, and race) from unconstrained face images.
Since the number of young subjects (\eg in the age group $0$--$20$) in the LFW database is very small (only $209$ subjects among the $5,749$ subjects according to the labels provided by MTurk workers), the LFW database was extended by collecting $2,466$ unconstrained face images of subjects in the age range $0$--$20$ years using Google Images search service.
Specifically, we first used the keywords such as ``baby", ``kid", and ``teenager" to find about $5,000$ images of interest from Google Images. The Viola-Jones \cite{ViolaIJCV04} face detector was then applied to generate a set of candidate faces.
Finally, we manually removed false face detections as well as most of the subjects that appeared to be older than $20$.
The extended LFW database (LFW+) contains $15,699$ unconstrained face images of about $8,000$ subjects.
For each face image, three MTurk workers were asked to provide their estimates of age, gender, and race.
The apparent age is determined as the average of the three estimates, and the gender and race are determined by the majority vote rule.
Results on LFW+ are reported with a five-fold, subject-exclusive cross-validation protocol.

\begin{figure}[t]
\begin{center}
\includegraphics[width=1\linewidth,height=1.15\linewidth]{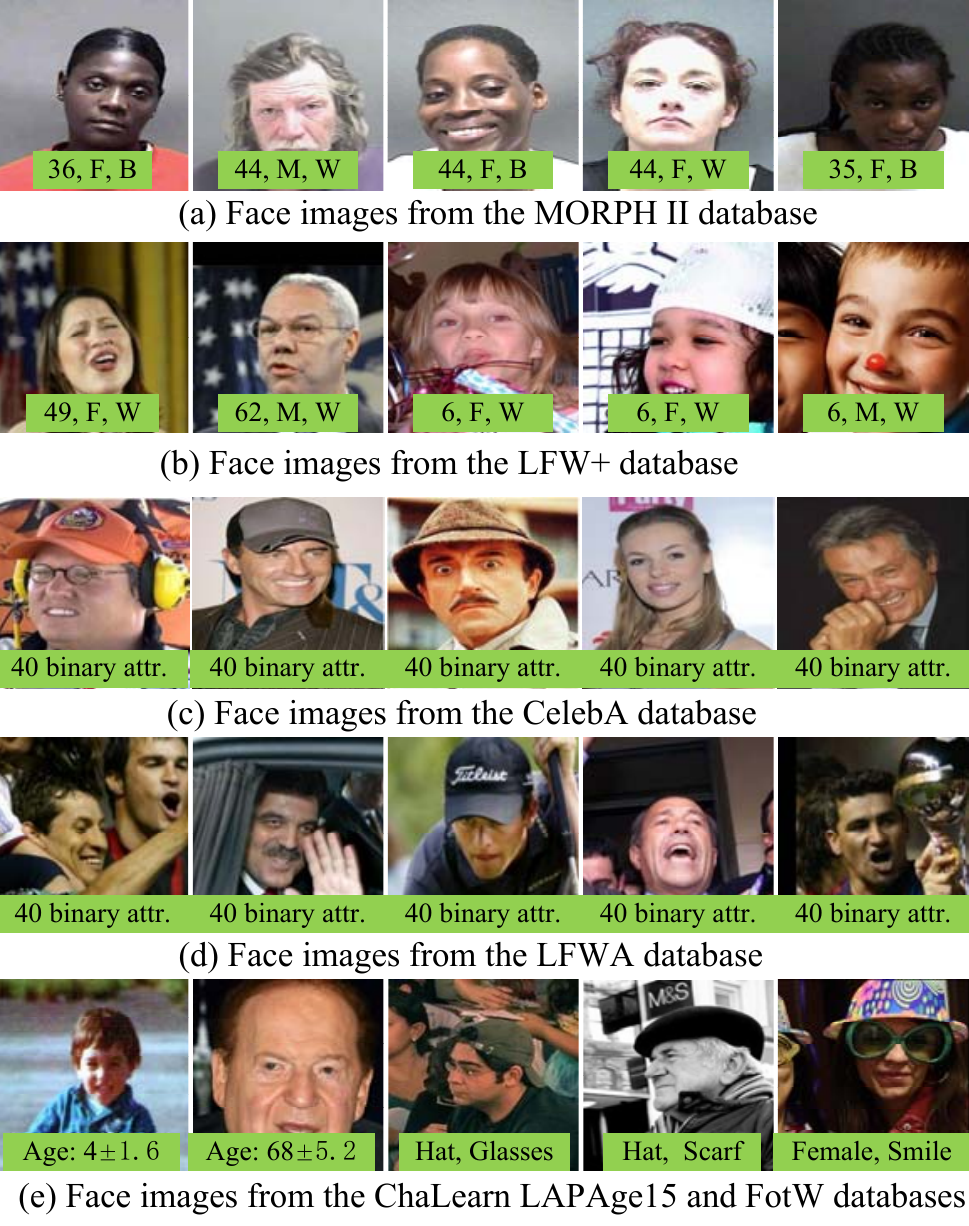}
\end{center}
\vspace{-4mm}
\caption[]{Examples of face images with nominal and ordinal attributes from (a) MORPH database (total of $78K$ face images) \cite{RicanekFG06}, and (b) LFW+ database (total of $15K$ face images); face images with $40$ binary attributes from (c) CelebA database (total of $200K$ face images) \cite{LiuICCV15}, and (d) LFWA database (total of $13K$ images) \cite{LiuICCV15}; and face images from (e) ChaLearn LAPAge2015 and FotW databases (total of $4K$ and $30K$ face images) \cite{EscaleraICCVW15}. M/F and B/W in (a--b) denote the gender (male, female) and race (black, white) information, respectively.}
\label{fig:DBExamples}
\vspace{-6mm}
\end{figure}

These databases can be divided into three group based on the type of annotation method used: (i) databases with nominal and ordinal attributes (MORPH II and LFW+), (ii) databases with binary attributes (CelebA, LFWA and FotW), and (iii) databases with a single attribute (LAPAge2015).
Example face images from the six databases are shown in Fig. \ref{fig:DBExamples}.
We can see that except for the MORPH II database, the other five databases mainly contain unconstrained face images.
Evaluations of attribute estimation on such databases could provide insights of the system's performance under real application scenarios.
In addition, we also evaluate the generalization ability of the proposed approach under \emph{cross-database testing} scenarios\footnotemark[6]\footnotetext[6]{In a cross-database testing, the attribute estimation method is trained on one face database, and tested on a different face database.}.

\subsection{Experimental Settings}
\label{Sec.ExpSetting}

For all the face images, we perform face and landmark detection using an open source SeetaFaceEngine\footnotemark[7]\footnotetext[7]{\url{https://github.com/seetaface}}, and normalize the face images into $256\times 256 \times 3$ (height $\times$ width $\times$ channels) based on five facial landmarks (\ie  two eye centers, nose tip, and two mouth corners).
Unless otherwise stated, we pre-train our DMTL network on the CASIA-WebFace database \cite{YiarXiv2014}, and then fine-tune this model on the training set of each individual database.
We use a base learning rate of $0.0001$, and reduce the learning rate to $10\%$ every $100,000$ iterations.
All the training and testing (except for our prototype system) are performed on a Nvidia Titan X GPU.
For the baseline methods used in Sections \ref{Sec.HeterFA}, \ref{Sec.BinaryFA}, and \ref{Sec.SingleFA} for which the code is not available in the public domain, we directly report the results in their publications.

There is no constraint in the network architecture for the shared feature learning in our DMTL.
We tried two networks (AlexNet \cite{KrizhevskyNIPS12} and GoogLeNet \cite{szegedy2015going}) with varying depths for attribute estimation on CelebA.
The average accuracies of all the $40$ attributes by AlexNet and GoogLeNet are $91.98\%$ and $92.05\%$, respectively.
The performance difference is minor, but AlexNet is much faster.
Therefore, we choose to use AlexNet (with a few modifications as described in Section \ref{Sec.ImpDetails}) for shared feature learning in our DMTL.


\begin{table}[t]
\caption{Estimation accuracies of the three heterogeneous attributes (age, gender, and race) on the MORPH II and LFW+ databases (in \%).}
\label{tab:HeterFARes}
\vspace{-3mm}
\renewcommand{\arraystretch}{1.1}
\centering
\begin{threeparttable}
\begin{tabular}{>{\raggedright\arraybackslash}p{1.3cm}|>{\centering\arraybackslash}p{0.9cm}>{\centering\arraybackslash}p{0.6cm}>{\centering\arraybackslash}p{0.6cm}|>{\centering\arraybackslash}p{0.9cm}>{\centering\arraybackslash}p{0.6cm}>{\centering\arraybackslash}p{0.6cm}}
\toprule
\multirow{2}{*}{\textbf{Approach}} & \multicolumn{3}{c}{\textbf{MORPH II}} & \multicolumn{3}{c}{\textbf{LFW+}} \\
\cline{2-7}
                                   & Age$^2$ & Gender & Race                   & Age$^2$ & Gender & Race \\
\hline
Guo and Mu \cite{GuoIVC14} & \multirow{2}{*}{3.92/70.0} & \multirow{2}{*}{98.5} & \multirow{2}{*}{99.0} & \multirow{2}{*}{NA} & \multirow{2}{*}{NA} & \multirow{2}{*}{NA} \\
Yi \etal \cite{YiACCV14}   & \multirow{2}{*}{3.63/NA} & \multirow{2}{*}{98.0} & \multirow{2}{*}{99.1} & \multirow{2}{*}{NA} & \multirow{2}{*}{NA} & \multirow{2}{*}{NA} \\
DIF \cite{HanTPAMI15}      & 3.8/75.0 & 97.6 & 99.1$^3$ & 7.8/42.5$^4$ & 94$^4$ & 90$^{3,4}$ \\
DEX \cite{RotheIJCV16}     & 3.25/NA  & NA   & NA & NA & NA & NA \\
DEX \cite{RotheIJCV16}$^1$ & 2.68/NA  & NA   & NA & NA & NA & NA\\
Proposed                   & 3.0/85.3 & 98.0 & 98.6 & 4.5/75.0 & 96.7 & 94.9 \\
\bottomrule
\end{tabular}
$^1$The IMDB-WIKI database \cite{RotheIJCV16} was used for network pre-training.
$^2$Age estimation results are reported in terms of both mean absolute error (MAE) and the accuracy with a 5-year absolute error.
$^3$Only two race classes (White vs. other) were used in \cite{HanTPAMI15}, but the proposed approach used three classes (Black, White, and Other).
$^4$Only the frontal face images in LFW were used in \cite{HanTPAMI15}.
\end{threeparttable}
\vspace{-4mm}
\end{table}

\subsection{Nominal and Ordinal Face Attributes}
\label{Sec.HeterFA}


The MORPH II and LFW+ databases, which contain age, gender, and race annotations, represent the scenario with heterogeneous attributes of nominal and ordinal.
Table \ref{tab:HeterFARes} lists the performance by the proposed approach and the state-of-the-art methods \cite{GuoIVC14,YiACCV14,HanTPAMI15} on the MORPH II and LFW+ databases.
Methods in \cite{GuoIVC14} and \cite{YiACCV14} provided a joint estimation of three face attributes, but both methods used multi-label regression.
Since the performance of \cite{GuoIVC14} and \cite{YiACCV14} is not available on the LFW+ database, we only compare the proposed approach with \cite{GuoIVC14} and \cite{YiACCV14} on the MORPH II database.
While our results on gender and race estimations are comparable with \cite{GuoIVC14} and \cite{YiACCV14}, the proposed approach performs much better than \cite{GuoIVC14} and \cite{YiACCV14} on the more challenging age estimation task ($3.0$ years MAE by the proposed approach vs. $3.92$ and $3.63$ years MAE by \cite{GuoIVC14} and \cite{YiACCV14}, respectively).
The possible reason is that multi-label learning in \cite{GuoIVC14} and \cite{YiACCV14} utilizes the same features for estimating different attributes, which may not be optimal.
By contrast, the subnetworks in our approach can fine-tune the shared features to obtain better feature representation for individual attributes.

Another baseline we considered is DEX \cite{RotheIJCV16}, which is not a multi-attribute estimation method, but reported the best known age estimation accuracy on MORPH II ($3.25$ years MAE).
Under the same settings, our approach performs better than DEX \cite{RotheIJCV16}, which suggests that by leveraging attribute correlations via MTL, our simple network can be as effective as a very deep VGG-16 network.
This also indicates that MTL could be a better choice than STL when multiple face attributes need to be jointly estimated.

\begin{figure}[t]
\begin{center}
\includegraphics[width=1\linewidth,height=.9\linewidth]{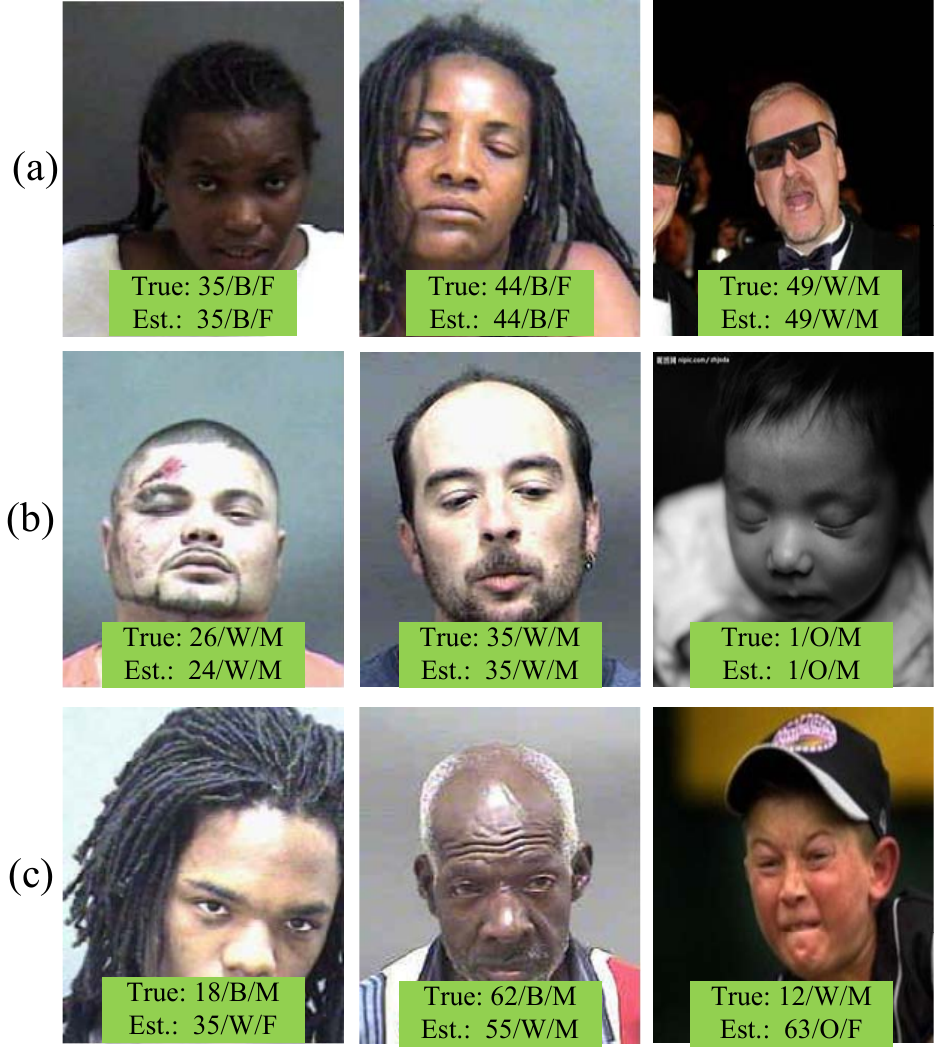}
\end{center}
\vspace{-4mm}
\caption[]{Examples of (a,b) good and (c) poor estimates for age, gender, and race by the proposed approach on the MORPH II and LFW+ databases. `m/n/l' denotes the age/race/gender information of each image, with `M/F' denoting male/female, and `W/O' denoting white/other, respectively.}
\label{Fig.ResExample_MORPH_LFW}
\vspace{-6mm}
\end{figure}


\begin{table*}[t]
\caption{Attribute estimation accuracies (in \%) for the $40$ binary attributes (see Table \ref{tab:CelebAttrInx}) on the CelebA and LFWA databases by the proposed approach and the state-of-the-art methods \cite{KumarECCV08,ZhangCVPR14,LiuICCV15,ZhongICB16,HandArXiv2016}. The average accuracies of \cite{KumarECCV08},\cite{ZhangCVPR14},\cite{LiuICCV15},\cite{ZhongICB16}, \cite{HandArXiv2016}, and the proposed approach are 81\%, 85\%, 87\%, 86.6\%, 91\% and 93\%, respectively, on CelebA, and 74\%, 81\%, 84\%, 84.7\%, 86.0\%, and 86.0\%, respectively, on LFWA.}
\vspace{-3mm}
\label{tab:BinaryFA}
\footnotesize
\centering
\begin{tabular}[width=1\linewidth]{cl|cccccccccccccccccccc}
\toprule
 & \multirow{2}{*}{\textbf{Approach}} &  \multicolumn{20}{c}{\textbf{Attribute index}} \\
 & & 1 & 2 & 3 & 4 & 5 & 6 & 7 & 8 & 9 & 10 & 11 & 12 & 13 & 14 & 15 & 16 & 17 & 18 & 19 & 20 \\
\hline

\multirow{5}{*}{\rotatebox{90}{\textbf{CelebA}}} & FaceTracker \cite{KumarECCV08} & 85 & 76 & 80 & 78 & 76 & 89 & 88 & 70 & 80 & 60 & 90 & 64 & 74 & 81 & 86 & 88 & 98 & 93 & 85 & 84 \\
& PANDA \cite{ZhangCVPR14}    &  88 & 78 & 86 & 81 & 79 & 96 & 92 & 85 & 93 & 77 & 94 & 67 & 75 & 86 & 86 & 88 & 98 & 93 & 90 & 86 \\
& LNets+ANet \cite{LiuICCV15} &  91 & 79 & 90 & 81 & 79 & 98 & 95 & 88 & 95 & 80 & 97 & 68 & 78 & 84 & 91 & 92 & 99 & 95 & 90 & 87 \\
& CTS-CNN \cite{ZhongICB16}   &  89 & 83 & 87 & 82 & 79 & 96 & 94 & 87 & 93 & 79 & 95 & 70 & 79 & 87 & 88 & 89 & 99 & 94 & 91 & 87\\
& MCNN-AUX \cite{HandArXiv2016} & \textbf{95} & 83 & \textbf{93} & 83 & 85 & \textbf{99} & \textbf{96} & \textbf{90} & \textbf{96} & 89 & \textbf{98} & 71 & 85 & \textbf{96} & 96 & 96 & \textbf{100} & 97 & \textbf{92} & \textbf{88} \\
& Proposed &  \textbf{95} & \textbf{86} & 85 & \textbf{85} & \textbf{99} & \textbf{99} & \textbf{96} & 85 & 91 & \textbf{96} & 96 & \textbf{88} & \textbf{92} & \textbf{96} & \textbf{97} & \textbf{99} & 99 & \textbf{98} & \textbf{92} & \textbf{88} \\
\hline

\multirow{5}{*}{\rotatebox{90}{\textbf{LFWA}}} & FaceTracker \cite{KumarECCV08} &  70 & 67 & 67 & 71 & 65 & 77 & 72 & 76 & 88 & 62 & 78 & 68 & 73 & 73 & 67 & 70 & 90 & 69 & 88 & 77 \\
& PANDA \cite{ZhangCVPR14} &  \textbf{84} & 79 & 79 & 81 & 80 & 84 & 84 & 87 & 94 & 74 & 81 & 73 & 79 & 74 & 69 & 75 & 89 & 75 & 93 & 86 \\
& LNets+ANet \cite{LiuICCV15} &  \textbf{84} & 82 & 82 & 83 & 83 & 88 & 88 & 90 & \textbf{97} & 77 & 84 & 75 & 81 & 74 & 73 & 78 & \textbf{95} & 78 & 95 & 88 \\
& CTS-CNN \cite{ZhongICB16} &  77 & 83 & 83 & 79 & 83 & 91 & \textbf{91} & 90 & \textbf{97} & 76 & 87 & 78 & 83 & \textbf{88} & 75 & 80 & 91 & 83 & 95 & 88 \\
& MCNN-AUX \cite{HandArXiv2016} & 77 & 82 & \textbf{85} & 80 & 83 & 92 & 90 & \textbf{93} & \textbf{97} & 81 & \textbf{89} & 79 & \textbf{85} & 85 & 77 & 82 & 91 & 83 & \textbf{96} & 88 \\
& Proposed &  80 & \textbf{86} & 82 & \textbf{84} & \textbf{92} & \textbf{93} & 77 & 83 & 92 & \textbf{97} & \textbf{89} & \textbf{81} & 80 & 75 & \textbf{78} & \textbf{92} & 86 & \textbf{88} & 95 & \textbf{89} \\
\midrule
\midrule
 & \multirow{2}{*}{\textbf{Approach}} &  \multicolumn{20}{c}{\textbf{Attribute index}} \\
 & & 21 & 22 & 23 & 24 & 25 & 26 & 27 & 28 & 29 & 30 & 31 & 32 & 33 & 34 & 35 & 36 & 37 & 38 & 39 & 40 \\
 \hline
\multirow{5}{*}{\rotatebox{90}{\textbf{CelebA}}} & FaceTracker \cite{KumarECCV08} &  91 & 87 & 91 & 82 & 90 & 64 & 83 & 68 & 76 & 84 & 94 & 89 & 63 & 73 & 73 & 89 & 89 & 68 & 86 & 80 \\
& PANDA \cite{ZhangCVPR14}    & 97 & 93 & 93 & 84 & 93 & 65 & 91 & 71 & 85 & 87 & 93 & 92 & 69 & 77 & 78 & 96 & 93 & 67 & 91 & 84 \\
& LNets+ANet \cite{LiuICCV15} & 98 & 92 & 95 & 81 & 95 & 66 & 91 & 72 & 89 & 90 & 96 & 92 & 73 & 80 & 82 & \textbf{99} & 93 & 71 & 93 & 87 \\
& CTS-CNN \cite{ZhongICB16}   & \textbf{99} & 92 & 93 & 78 & 94 & 67 & 85 & 73 & 87 & 88 & 95 & 92 & 73 & 79 & 82 & 96 & 93 & 73 & 91 & 86 \\
& MCNN-AUX \cite{HandArXiv2016} & 98 & \textbf{94} & \textbf{97} & 87 & 96 & 76 & \textbf{97} & 77 & \textbf{94} & 95 & \textbf{98} & 93 & 84 & 84 & 90 & \textbf{99} & \textbf{94} & 87 & \textbf{97} & 88 \\
& Proposed                    & 98 & \textbf{94} & \textbf{97} & \textbf{90} & \textbf{97} & \textbf{78} & \textbf{97} & \textbf{78} & \textbf{94} & \textbf{96} & \textbf{98} & \textbf{94} & \textbf{85} & \textbf{87} & \textbf{91} & \textbf{99} & 93 & \textbf{89} & \textbf{97} & \textbf{90} \\
\hline

\multirow{5}{*}{\rotatebox{90}{\textbf{LFWA}}} & FaceTracker \cite{KumarECCV08} & 84 & 77 & 83 & 73 & 69 & 66 & 70 & 74 & 63 & 70 & 71 & 78 & 67 & 62 & 88 & 75 & 87 & 81 & 71 & 80 \\
& PANDA \cite{ZhangCVPR14}    & 92 & 78 & 87 & 73 & 75 & 72 & 84 & 76 & 84 & 73 & 76 & 89 & 73 & 75 & 92 & 82 & 93 & 86 & 79 & 82 \\
& LNets+ANet \cite{LiuICCV15} & \textbf{94} & 82 & 92 & 81 & 79 & 74 & 84 & 80 & 85 & 78 & 77 & 91 & 76 & 76 & 94 & 88 & \textbf{95} & 88 & 79 & 86 \\
& CTS-CNN \cite{ZhongICB16}   & \textbf{94} & 81 & 94 & 81 & 80 & 75 & 73 & 83 & \textbf{86} & 82 & 82 & 90 & 77 & 77 & 94 & 90 & \textbf{95} & 90 & \textbf{81} & 86 \\
& MCNN-AUX \cite{HandArXiv2016} & \textbf{94} & 84 & 93 & \textbf{83} & \textbf{82} & \textbf{77} & \textbf{93} & \textbf{84} & \textbf{86} & \textbf{88} & \textbf{83} & \textbf{92} & \textbf{79} & \textbf{82} & \textbf{95} & 90 & \textbf{95} & 90 & \textbf{81} & 86\\
& Proposed                    & 93 & \textbf{86} & \textbf{95} & 82 & 81 & 75 & 91 & \textbf{84} & 85 & 86 & 80 & \textbf{92} & \textbf{79} & 80 & 94 & \textbf{92} & 93 & \textbf{91} & \textbf{81} & \textbf{87} \\
\bottomrule
\end{tabular}
\vspace{-5mm}
\end{table*}

Among the multi-attribute estimation methods, only DIF \cite{HanTPAMI15} reported their results on a subset of LFW with frontal face images.
On this frontal subset, DIF \cite{HanTPAMI15} achieved $42.5\%$ (@ 5-year AE), $94\%$, and $90\%$ accuracies for age, gender, and race estimations, respectively.
The proposed DMTL achieves $75.0\%$ (@ 5-year AE), $96.7\%$, and $94.9\%$ accuracies for age, gender, and race estimations, on the much larger LFW+ database with unconstrained face images.

Examples of correct and incorrect age, gender, and race estimates by the proposed approach on the MORPH II and LFW+ databases are shown in Figure \ref{Fig.ResExample_MORPH_LFW}.
We find that the proposed approach is quite robust to pose and illumination variations. However, we also notice that the small number of young and old subjects in both the MORPH II and LFW+ databases can make the age and race estimation difficult.

\subsection{Binary Face Attributes}
\label{Sec.BinaryFA}

In practice, it is relatively easy to annotate the presence of each attribute (binary attribute) than fine-grained annotations (\eg nominal and ordinal).
The CelebA, LFWA and FotW databases represent the scenario of joint estimation for multiple binary attributes.
Binary attributes could be heterogeneous in terms of holistic vs. local (\eg in CelebA and LFWA), but no longer heterogeneous in terms of nominal vs. ordinal.
Therefore, we can handle binary attributes through holistic and local subnetworks with the same loss.
Specifically, for the CelebA and LFWA databases, we use one holistic nominal subnetwork (for attributes: \#4, 14, 15, 19, 21, 26, 27, 32, and 40 in Table \ref{tab:CelebAttrInx}) and seven local nominal subnetworks (subnet1 for attributes \#6, 7, 8, 9, 10, 11, 29, 33, 34, 36; subnet2 for attributes \#2, 3, 5, 17, 24; subnet3 for attributes: \#13, 28; subnet4 for attributes: \#20, 30, 31, 35; subnet5 for attributes: \#1, 12, 22, 23, 37; subnet6 for attributes: \#16, 18, 25; subnet7 for attributes: \#38, 39).

\begin{figure}[t]
\begin{center}
\includegraphics[width=1\linewidth,height=.9\linewidth]{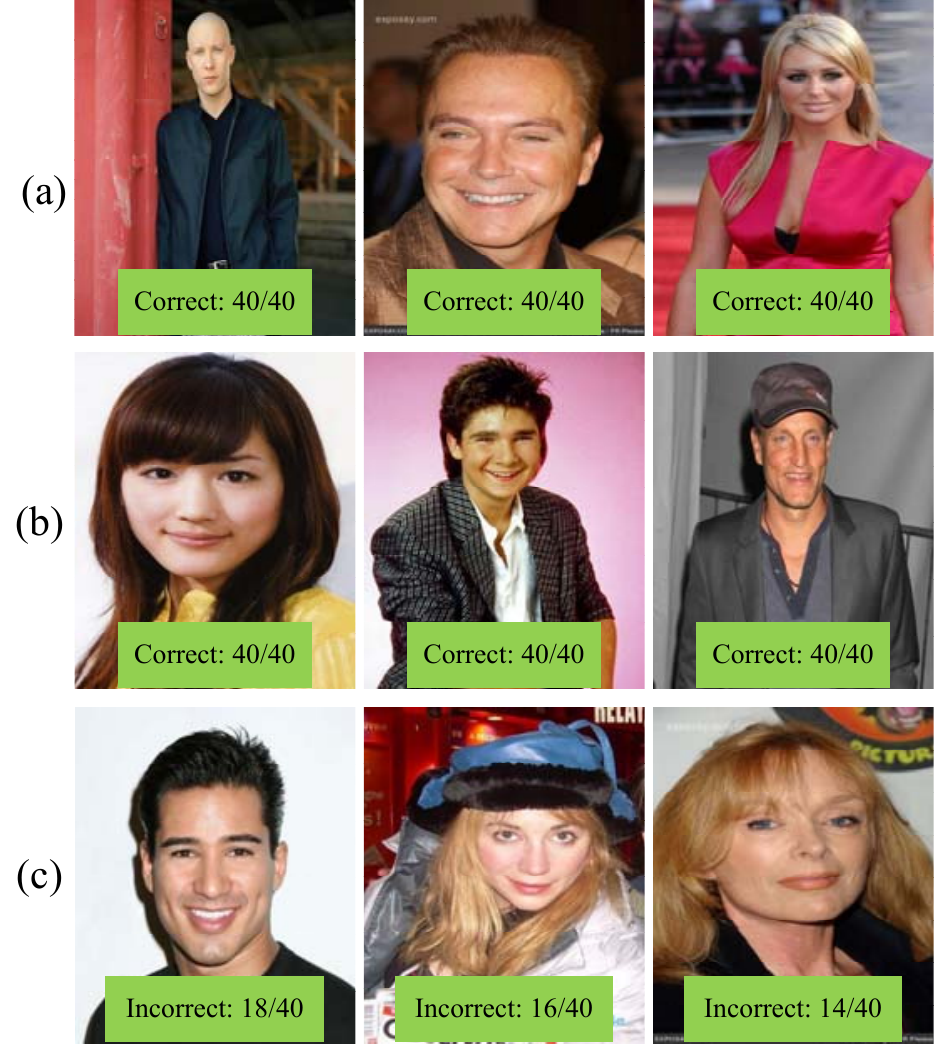}
\end{center}
\vspace{-4mm}
\caption[]{Examples of (a,b) good and (c) poor estimates for the $40$ binary face attributes by the proposed approach on the CelebA databases.  `m/n' denotes (the number of correct estimates)/(total number of attributes) for each face image.}
\label{Fig.ResExample_CelebA}
\vspace{-6mm}
\end{figure}

The results on CelebA and LFWA by the proposed approach and several state-of-the-art methods \cite{LiuICCV15,ZhongICB16,HandArXiv2016,ZhangCVPR14,KumarECCV08} are reported in Table \ref{tab:BinaryFA}.
The proposed approach outperforms \cite{LiuICCV15,ZhongICB16,ZhangCVPR14,KumarECCV08} for most of the $40$ face attributes on both the CelebA and LFWA databases.
Comparisons with \cite{KumarECCV08,LiuICCV15}, which used per attribute SVM classifiers, show superior performance of the proposed DMTL in jointly estimating multiple attributes.
Our approach achieves similar accuracies to MCNN-AUX \cite{HandArXiv2016} on LFWA.
The possible reason is that both methods tend to show overfitting on such a small training set of LFWA ($6K$ images), leading to unsatisfactory results on the testing set.
Given a larger training set such as CelebA ($160K$ images), both methods are improved, but our method performs better than \cite{HandArXiv2016}.
Figure \ref{Fig.ResExample_CelebA} shows examples of good and poor attribute estimates by our approach on the CelebA database.
Some of the poor estimates by the proposed approach are due to the inconsistencies in the provided attributes.
For example, the first image in Fig. \ref{Fig.ResExample_CelebA} (c) was labeled with both attribute \#1 `5 o'Clock Shadow' and attribute \#25 `No Beard'.

\begin{figure}[!t]
\includegraphics[width=0.49\textwidth,height=.26\textwidth]{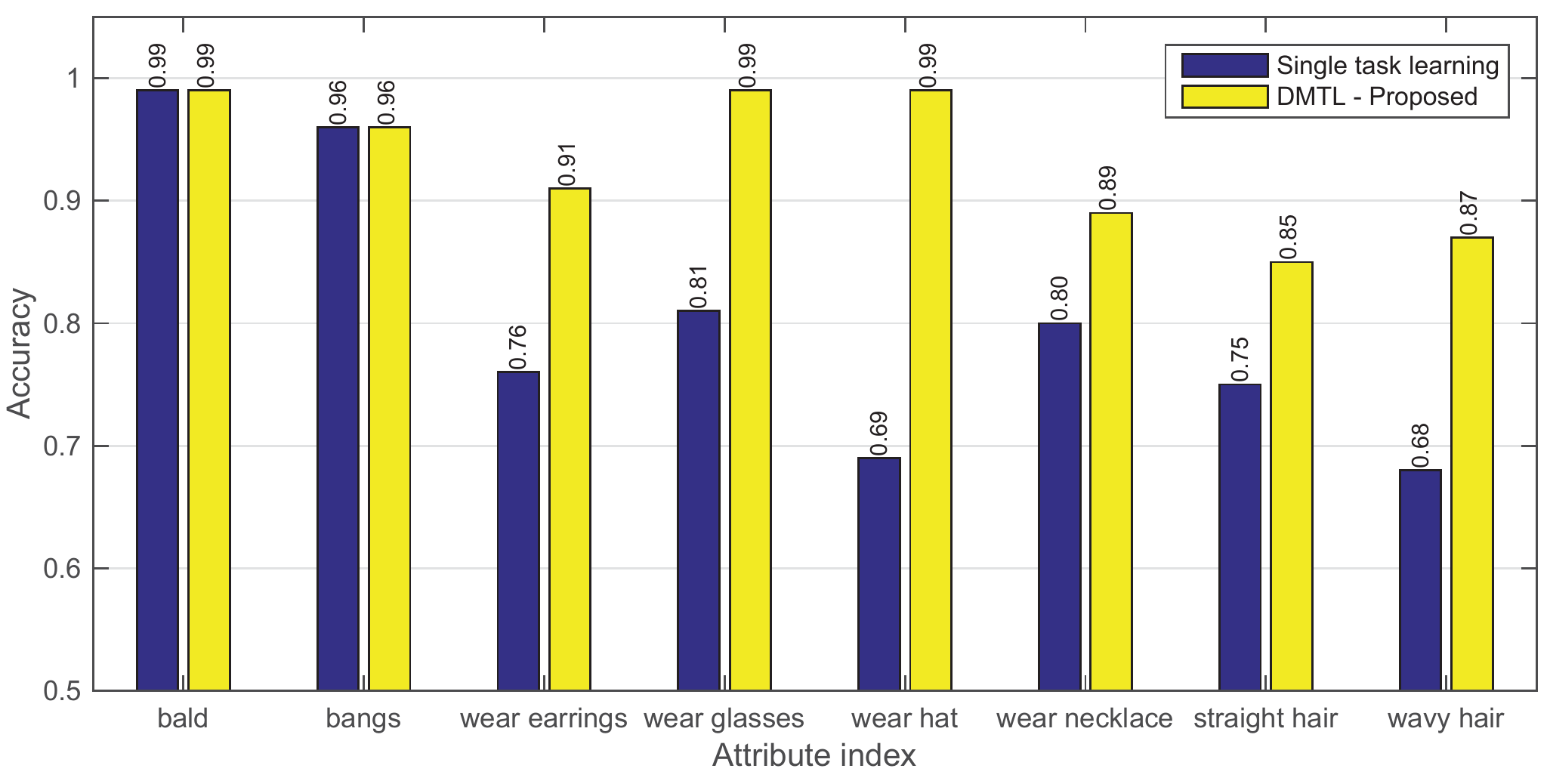}
\vspace{-8mm}
\caption[MTLvsSTL]{Attribute estimation accuracies by the proposed DMTL approach and the baseline single-task learning (STL) method for eight common attributes from the CelebA database. On average, MTL works much better than STL using networks with a similar depth.}
\label{Fig.CelebA10_DMTLVSSTL}
\end{figure}

We also provide the results by STL, \ie training a separate AlexNet model for each face attribute.
Since there are up to $40$ face attributes in CelebA, we simply chose eight common attributes.
Figure \ref{Fig.CelebA10_DMTLVSSTL} shows that while STL may work well for a few attributes, overall the proposed DMTL performs much better than STL.
It is not clear to what degree the attribute correlations were utilized in the published methods, but we checked the incorrect estimation results for attribute \#38 (`WearNecklace') by our approach, and find that the number of males (attribute \#21) satisfying this attribute is very small.
This makes sense because males wear necklace much less often than females do.

For the two FotW datasets, since there is no clear attribute heterogeneity, either nominal vs. ordinal or holistic vs. local, we simply use a nominal subnetwork in our DMTL.
Results by our approach and the state-of-the-art methods (reported in \cite{EscaleraCVPRW16}) for accessory classification, and smile and gender classification on FotW are shown in Table \ref{tab:FotWAccuracy}.
Our approach achieves an average accuracy of $94.0\%$ for accessory classification, which is better than the best result ($93.5\%$) by SIAT MMLAB \cite{ZhangCVPRW16}.
For smile and gender classification, our approach achieves an average accuracy of $86.1\%$, which is lower than the top-2 methods (SIAT MMLAB \cite{ZhangCVPRW16} and IVA NLPR \cite{LiCVPRW16}) reported in \cite{EscaleraCVPRW16}.
However, while methods in \cite{ZhangCVPRW16,LiCVPRW16} used very deep networks like VGG \cite{SimonyanArXiv2014}, our approach only uses a network with complexity similar to AlexNet.

These results indicate that our DMTL can make use of attribute correlations to achieve better attribute estimation results.
In addition, our DMTL is effective in handling attribute heterogeneities, \eg nominal vs. ordinal and holistic vs. local, by using different number and different types of subnetworks for category-specific feature learning.

\subsection{Single Face Attribute}
\label{Sec.SingleFA}

Some application scenarios may require the estimate of a single attribute, \eg age estimation used for preventing minors from purchasing alcohol or cigarette from camera-enabled vending machines.\footnotemark[8]\footnotetext[8]{\url{http://newsfeed.time.com/2011/12/27/scram-kids-new-vending-machine-dispenses-pudding-to-adults-only}}
The LAPAge2015 database represents such a scenario with age estimation from unconstrained face images.
Following \cite{RotheIJCV16}, we train our DMTL network without and with pre-training on the IMDB-WIKI database\footnotemark[9]\footnotetext[9]{\url{https://data.vision.ee.ethz.ch/cvl/rrothe/imdb-wiki}}.
Both the MAE and $\epsilon$-error ($\varepsilon=1-\exp(-\frac{(y-\mu)^2}{2\sigma^2})$) are used to measure the performance.
When the proposed DMTL network is trained from scratch using only the training set of the LAPAge2015 database, it achieves an $\epsilon$-error of $0.449$, and $5.2$ years MAE.
This result is comparable to the $8$-th best method among all the $115$ participants of LAPAge2015 \cite{EscaleraICCVW15}.
If we pre-train our DMTL approach using the IMDB-WIKI database, and then fine-tune the model on the training set of the LAPAge2015 database, the proposed approach achieves an $\epsilon$-error of $0.289$.
This result is comparable to the best age estimation result (an $\epsilon$-error of $0.265$) on LAPAge2015, which was reported by DEX in \cite{RotheIJCV16}.
However, while DEX \cite{RotheIJCV16} is an ensemble of $20$ VGG-16 networks, the proposed approach is a single network with complexity similar to AlexNet.

\begin{figure}[!t]
\begin{center}
\includegraphics[width=1\linewidth,height=0.9\linewidth]{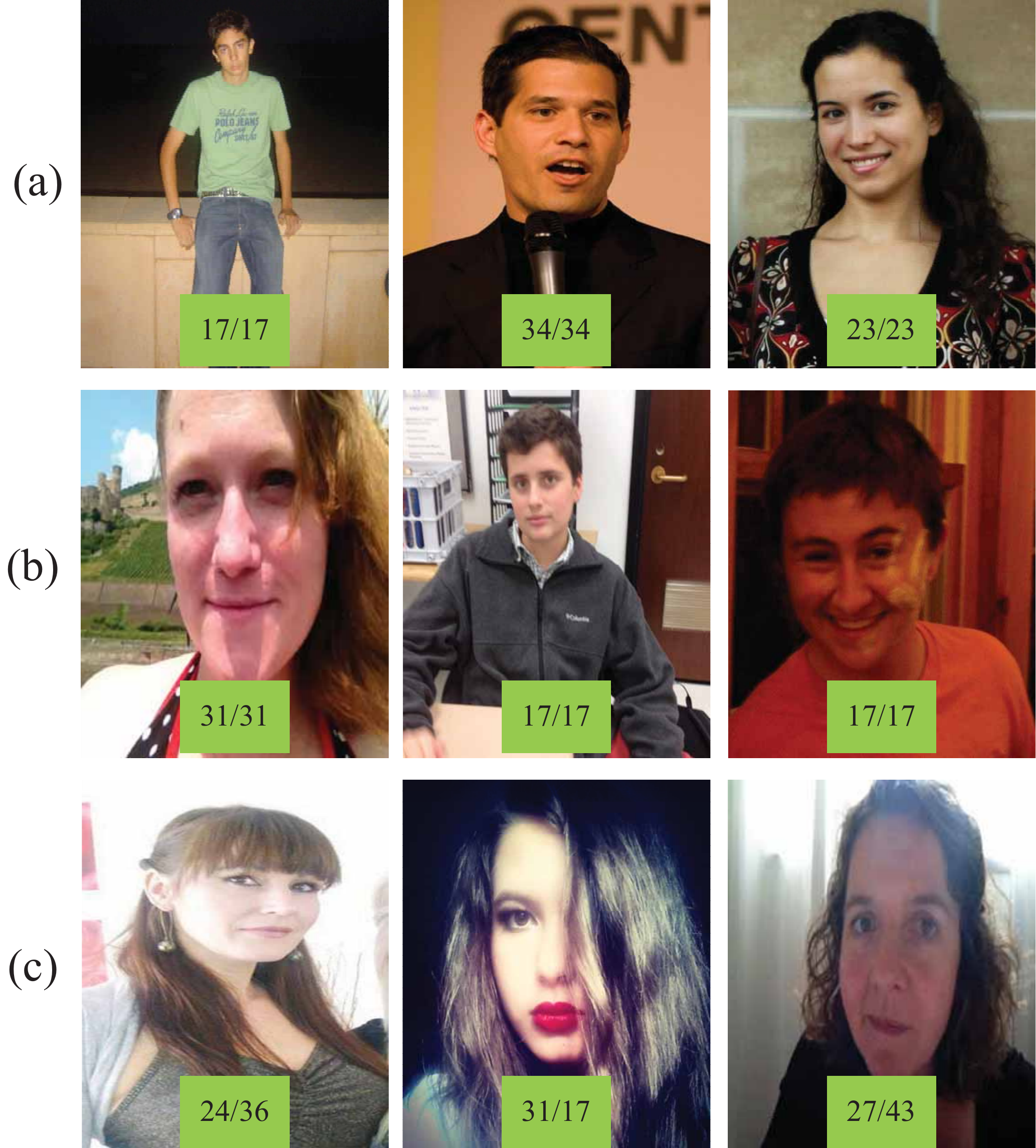}
\end{center}
\vspace{-4mm}
\caption[]{Examples of (a,b) good and (c) poor age estimations by the proposed approach on the LAPAge2015 database. `m/n' denotes the (estimated age)/(ground-truth apparent age) respectively, for each face image.}
\label{Fig.ResExample_LAP}
\vspace{-6mm}
\end{figure}

Figure \ref{Fig.ResExample_LAP} shows examples of good and poor age estimates by our approach for age estimation on the LAPAge2015 database.
Loss of face details due to overexposure of the image is responsible for some poor age estimates (see Fig. \ref{Fig.ResExample_LAP} (c)).

\begin{table}[t]
\caption{Accuracies (in \%) of the proposed approach and the state-of-the-art methods (reported in \cite{EscaleraCVPRW16}) for (a) accessory classification, and (b) smile and gender classification on the FotW datasets.}
\label{tab:FotWAccuracy}
\vspace{-3mm}
\footnotesize
\centering
\begin{tabular}[width=1\linewidth]{>{\raggedright\arraybackslash}p{1.15cm}>{\raggedright\arraybackslash}p{0.25cm}>{\raggedright\arraybackslash}p{0.5cm}>{\raggedright\arraybackslash}p{0.45cm}>{\raggedright\arraybackslash}p{0.45cm}>{\raggedright\arraybackslash}p{0.45cm}>{\raggedright\arraybackslash}p{0.45cm}>{\raggedright\arraybackslash}p{0.45cm}>{\raggedright\arraybackslash}p{0.45cm}}
\toprule
\textbf{Method} & \tiny{\textbf{Hat}} & \tiny{\textbf{Headband}}	& \tiny{\textbf{Glasses}} & \tiny{\textbf{Earrings}} & \tiny{\textbf{Necklace}} & \tiny{\textbf{Tie}} & \tiny{\textbf{Scarf}} & \tiny{\textbf{Avg.}} \\
\hline
SIAT MMLAB & \multirow{2}{*}{94.7} & \multirow{2}{*}{94.9} & \multirow{2}{*}{94.7} & \multirow{2}{*}{91.0} & \multirow{2}{*}{88.2} & \multirow{2}{*}{97.3} & \multirow{2}{*}{93.7} & \multirow{2}{*}{93.5}\\
IVA NLPR & \multirow{2}{*}{92.2} & \multirow{2}{*}{95.1} & \multirow{2}{*}{93.9} & \multirow{2}{*}{85.3} & \multirow{2}{*}{87.4} & \multirow{2}{*}{96.1} & \multirow{2}{*}{94.0} & \multirow{2}{*}{92.0} \\
\textbf{Proposed}   & 94.7 & 96.1 & 96.1 & 89.1 & 89.5 & 97.4 & 95.1 & 94.0\\
\bottomrule
\end{tabular}

(a) FotW - accessory classification
\vspace{1mm}

\begin{tabular}[width=1\linewidth]{lccc}
\toprule
\textbf{Method} & \textbf{Smile} & \textbf{Gender}	& \textbf{Avg.} \\
\hline
SIAT MMLAB & 92.7 & 85.8 & 89.3 \\
IVA NLPR   & 91.5 & 82.5 & 87.0 \\
VISI.CRIM  & 90.2 & 82.1 & 86.1 \\
SMILELAB NEU & 90.0 & 81.5 & 85.7 \\
\textbf{Proposed}   & 84.9 & 87.3 & 86.1 \\
\bottomrule
\end{tabular}

(b) FotW - smile and gender classification
\vspace{-6mm}
\end{table}

\subsection{Generalization Ability}
\label{Sec.Generalizaion}

The data distribution in the system deployment environment can be different from that during model development.
We evaluate the generalization ability of the proposed approach with cross-database testing on the MORPH II, LFW+, CelebA, and LFWA databases.

Specifically, cross-database testing of age, gender, and race estimation between the MORPH II and LFW+ databases is performed by training our approach on LFW+ and testing it on MORPH II, and vice versa.
Similarly, cross-database testing of $40$ face attribute estimation is performed between the CelebA and LFWA databases.
The attribute estimation results with cross-database testing are shown in Table \ref{tab:CrossDBTest}.
As expected, cross-database testing performance is lower than intra-database testing.
But, we believe these accuracies (not reported in other published studies) are still quite good.
Image distribution (age, gender, race, pose, expression, occlusion, and illumination) differences between the MORPH II and LFW+ databases are responsible for the drop in performance.
For example, there are more males than females in the MORPH II ($84\%$) and LFW+ ($74\%$) databases, and the race distributions in MORPH II and LFW+ are significantly biased towards black ($75\%$) and white ($79\%$), respectively.
The reasons for the drop in performance of the cross-database testing between CelebA and LFWA are similar.
In addition, although both CelebA and LFWA contain face images of individuals such as celebrities, public figures, etc., face images in LFWA were selected by using the Viola-Jones face detector \cite{ViolaIJCV04}.
Thus, face images in LFWA have relatively small variations in pose, expression, occlusion, etc.
Finally, the LFWA database only contains $13,233$ face images, making it difficult to train a robust CNN model.

We also combine the age and race information in our LFW+ database with the $40$ attributes in the LFWA database, leading to a new LFWA database (LFWA+) with $42$ attributes.\footnotemark[10]\footnotetext[10]{The gender information is already provided with the LFWA database.}
Since LFW+ and LFWA were constructed independently, crowdsourcing methods used in the two databases could be different.
We evaluate the proposed approach using LFWA+ to see its effectiveness in handling both attribute heterogeneity and different annotation sources.
We used a five-fold, subject-exclusive cross-validation protocol.
The proposed approach using nominal and ordinal subnetworks achieves $4.8$ years MAE for age estimation, $91\%$ accuracy for race classification, and $83\%$ accuracy for the average of the other $40$ attributes.
Compared with the results on the separate LFW+ and LFWA databases (see Tables \ref{tab:HeterFARes} and \ref{tab:BinaryFA}), the accuracies on the combined LFWA+ database are slightly lower.
This experiment indicates that different sources of annotations may pose additional challenges to face attribute estimation, but the proposed approach still achieves quite good results in such a challenging scenario.

\begin{table}[!t]
\begin{center}
\caption{Cross-database testing accuracies (in \%) of the proposed approach using MORPH II and LFW+, as well as CelebA and LFWA.}
\label{tab:CrossDBTest}
\begin{threeparttable}
\begin{tabular}{ccccc}
\toprule
\multicolumn{2}{c}{\textbf{Database}} & \multicolumn{3}{c}{\textbf{Accuracy}} \\
Training & Testing & Age$^1$ & Gender & Race \\
\midrule
MORPH II & MORPH II & 3.0/85.3 & 98.0 & 98.6 \\
LFW+ & MORPH II     & 7.0/60.1 & 89.0 & 85.7     \\
\cline{3-5}
LFW+     & LFW+ & 4.5/75.0 & 96.7 & 94.9 \\
MORPH II & LFW+ & 9.4/52.6 & 77.4 & 70.5 \\
\hline
           &        & \multicolumn{3}{c}{\textbf{Avg. accuracy of 40 attributes}}\\
CelebA     & CelebA &  \multicolumn{3}{c}{93.0} \\
LFWA       & CelebA &  \multicolumn{3}{c}{70.2} \\
\cline{3-5}
LFWA     & LFWA   &  \multicolumn{3}{c}{86.0} \\
CelebA   & LFWA   &  \multicolumn{3}{c}{73.0} \\
\bottomrule
\end{tabular}
$^1$Age estimation results are reported using both mean absolute error (MAE) and the accuracy with a 5-year AE.
\end{threeparttable}
\end{center}
\vspace{-8mm}
\end{table}

\subsection{Computational Cost}
\label{Sec.ComputCost}


We summarize the computational cost of the proposed approach and several state-of-the-art methods on the MORPH II, CelebA, LFWA, and LAPAge2015 databases.
For feature learning and joint attribute estimation, the proposed approach takes $8$ms on a Titan X GPU, and $35$ms on an Intel Core I7 3.6 GHz CPU.
Only a few of the state-of-the-art methods reported their computational costs using machines with different GPUs and CPUs.
We still report their computational costs for reference in Table \ref{tab:ComputCost}.
Compared with the methods that reported computational cost on GPU, the proposed approach is much faster than state-of-the-art methods except for MS-CNN \cite{YiACCV14}. However, our approach works much better than \cite{YiACCV14} for age estimation on MORPH II.
Compared with the best method on LAPAge2015 (DEX \cite{RotheIJCV16}), the proposed approach is about $10$ times faster than a single VGG-16 model used in \cite{RotheIJCV16}.
For the computational cost on CPU, the proposed approach is faster than the rKCCA method in \cite{GuoIVC14} and MS-CNN in \cite{YiACCV14}.
A prototype implementation of the proposed approach is able to run in real-time (about $16$ fps) on the CPU (Intel Core I7 3.6 GHz) of a commodity desktop machine (see a demo at: {http://ddl.escience.cn/f/FOrq}), which suggests that our approach can be used in wide application scenarios.

\begin{table}[!t]
\begin{center}
\caption{Computational cost of different face attribute estimation methods.}
\label{tab:ComputCost}
\renewcommand{\arraystretch}{1.1}
\footnotesize
\begin{threeparttable}
\begin{tabular}{lccc}
\toprule
\textbf{Method} & \textbf{\tabincell{c}{Face \\Detection}} & \textbf{\tabincell{c}{Feature \\learning}} & \textbf{Prediction} \\
\midrule
\tabincell{l}{MS-CNN \cite{YiACCV14}} (GPU) & N/A & \multicolumn{2}{c}{$2$ms$^1$ } \\
\tabincell{l}{LNet+ANet \cite{LiuICCV15}} (GPU) & $35$ms & $14$ms & N/A\\
DEX \cite{RotheIJCV16} (GPU) & N/A & \multicolumn{2}{c}{$\sim75$ms$^2$ with VGG-16}\\
\textbf{Proposed} (GPU)       & $5$ms$^1$  & \multicolumn{2}{c}{$8$ms$^2$ }\\
\tabincell{l}{rKCCA \cite{GuoIVC14} (CPU)}  & N/A & N/A & $1,600$ms$^3$ \\
\tabincell{l}{MS-CNN \cite{YiACCV14} (CPU)} & N/A & \multicolumn{2}{c}{$200$ms$^4$ } \\
\textbf{Proposed} (CPU) & $25$ms$^5$ & \multicolumn{2}{c}{$35$ms$^5$ }\\
\bottomrule
\end{tabular}
$^{1,2,3,4,5}$The computational costs are profiled on a Tesla K20 GPU, Titan X GPU, Intel Core2 2.1 GHz CPU, Intel Core I3 2.4 GHz CPU, Intel Core I7 3.6 GHz CPU, respectively.

\end{threeparttable}
\end{center}
\vspace{-6mm}
\end{table}

\section{Conclusions}
\label{Sec.Summary}

This paper presents a deep multi-task learning approach for joint estimation of multiple face attributes.
Compared to the existing approaches, the proposed approach models both attribute correlation and attribute heterogeneity in a single network, allowing shared feature learning for all the attributes, and category-specific feature learning for heterogeneous attributes.
The LFW+ database was created by augmenting the LFW database with $2,466$ images of subjects in $0$-$20$ years of age.
This helps evaluate the proposed approach on a wider age range.
Our approach performs well on large and diverse databases (including MORPH II, LFW+, CelebA, LFWA, LAPAge2015, and FotW), which replicate several representative scenarios such as face databases with multiple heterogeneous attributes and a single attribute.
Generalization ability of the proposed approach is studied under the cross-database testing scenarios.
Experimental results show that the proposed approach generalizes well to the unseen scenarios.
The cross-database testing highlights the importance of training database in real-world face attribute estimation systems.
Additionally, the ambiguity of annotation for some attributes would be another issue that makes it difficult to learn efficient models. One possible solution to this issue could be integrating noisy label refining with deep multi-task learning.

\ifCLASSOPTIONcompsoc
  \section*{Acknowledgments}
\else
  \section*{Acknowledgment}
\fi
This research was partially supported by the National Basic Research Program of China (973 Program) (grant 2015CB351802), Natural Science Foundation of China (grant 61390511, 61672496, and 61650202), and CAS-INRIA JRPs (grant FER4HM).
S. Shan is the corresponding author.
%

\ifCLASSOPTIONcaptionsoff
  \newpage
\fi



\bibliographystyle{IEEEtran}
\bibliography{dmtl_reference}

\begin{biography}[{\includegraphics[width=1in,height=1.25in,clip,keepaspectratio]{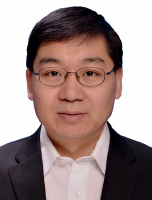}}]{Hu Han} is an Associate Professor of the Institute of Computing Technology (ICT), Chinese Academy of Sciences (CAS).
He received the B.S. degree from Shandong University, and the Ph.D. degree from ICT, CAS,  in 2005 and 2011, respectively, both in computer science.
He was a Research Associate in the Department of Computer Science and Engineering at Michigan State University, and a visiting researcher at Google in Mountain View from 2011 to 2015.
His research interests include computer vision, pattern recognition, and image processing, with applications to biometrics, forensics, law enforcement, and security systems. He is a member of the IEEE.
\end{biography}
\vspace{-0.7cm}
\begin{biography}[{\includegraphics[width=1in,height=1.25in,clip,keepaspectratio]{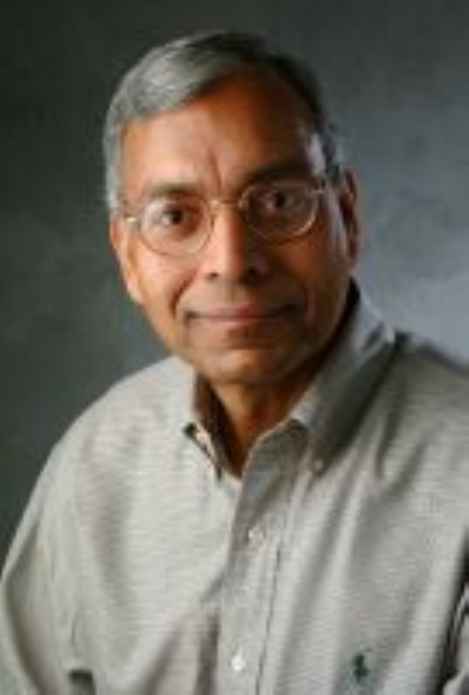}}]{Anil K. Jain} is a University Distinguished Professor in the Department of Computer Science and Engineering at Michigan State University. His research interests include pattern recognition and biometric authentication. He served as the editor-in-chief of the \textsc{IEEE Transactions on Pattern Analysis and Machine Intelligence} (1991-1994). He served as a member of the United States Defense Science Board and The National Academies committees on Whither Biometrics and Improvised Explosive Devices. He has received Fulbright, Guggenheim, Alexander von Humboldt, and IAPR King Sun Fu awards. He is a member of the National Academy of Engineering
and foreign fellow of the Indian National Academy of Engineering. He is a Fellow of the AAAS, ACM, IAPR, SPIE, and IEEE.
\end{biography}
\vspace{-0.7cm}
\begin{biography}[{\includegraphics[width=1in,height=1.2in,clip,keepaspectratio]{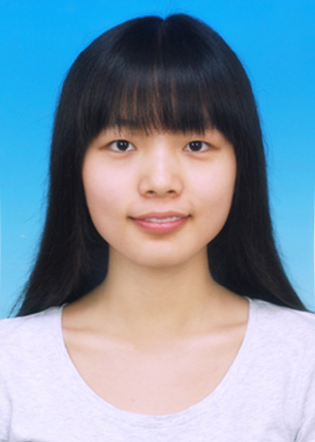}}]{Fang Wang} received the B.S. degree from Tianjin University in 2014,
and the M.S. degree from ICT, CAS in 2017. Her research interests include computer vision and pattern recognition.
\end{biography}
\begin{biography}[{\includegraphics[width=1in,height=1.25in,clip,keepaspectratio]{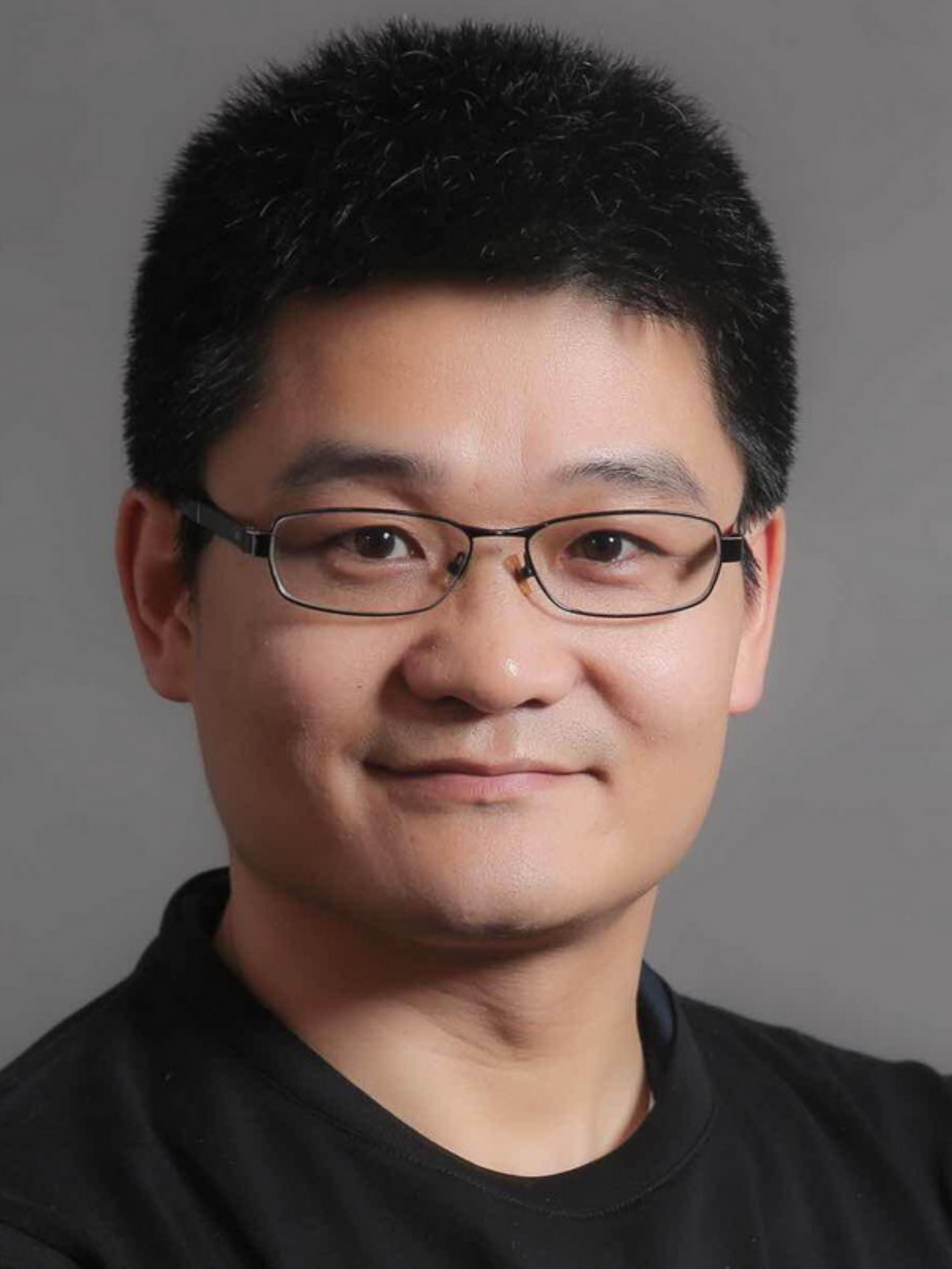}}]{Shiguang Shan} is a Professor of
ICT, CAS, and the Deputy Director with the Key Laboratory of Intelligent Information Processing, CAS.
His research interests cover computer vision, pattern recognition, and machine learning.
He has authored over 200 papers in refereed journals and proceedings in the areas of computer vision and pattern recognition.
He was a recipient of the China's State Natural Science Award in 2015, and the China’s State S\&T Progress Award in 2005 for his research work.
He has served as the Area Chair for many international conferences, including ICCV'11, ICPR'12, ACCV'12, FG'13, ICPR'14, and ACCV'16. He is an Associate Editor of several journals, including the \textsc{IEEE Transactions on Image Processing}, the Computer Vision and Image Understanding, the Neurocomputing, and the Pattern Recognition Letters.
He is a Senior Member of IEEE.
\end{biography}
\vspace{-0.7cm}
\begin{biography}[{\includegraphics[width=1in,height=1.25in,clip,keepaspectratio]{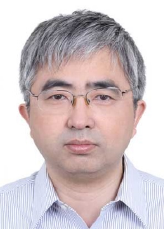}}]{Xilin Chen} is a Professor of
ICT, CAS.
He has authored one book and over 200 papers in refereed journals and proceedings in the areas of computer vision, pattern recognition, image processing, and multimodal interfaces.
He served as an Organizing Committee/Program Committee member for over 50 conferences. He was a recipient of several awards, including the China's State Natural Science Award in 2015, the China's State S\&T Progress Award
in 2000, 2003, 2005, and 2012 for his research work. He is currently an Associate Editor of the \textsc{IEEE Transactions on Multimedia}, a Leading Editor of the Journal of Computer Science and Technology, and an Associate Editor-in-Chief of the Chinese Journal of Computers.
He is a Fellow of the China Computer Federation (CCF), and IEEE.
\end{biography}

\vfill





\end{document}